\documentclass[10pt,twocolumn,letterpaper]{article}

\usepackage{cvpr}            % To produce the CAMERA-READY version

\usepackage[accsupp]{axessibility}

\usepackage[dvipsnames]{xcolor}

\definecolor{cvprblue}{rgb}{0.21,0.49,0.74}
\usepackage[pagebackref,breaklinks,colorlinks,citecolor=cvprblue]{hyperref}

\def\paperID{3733} % *** Enter the Paper ID here
\def\confName{CVPR}
\def\confYear{2024}

\title{KTPFormer: Kinematics and Trajectory Prior Knowledge-Enhanced Transformer for 3D Human Pose Estimation}

\author{Jihua Peng$^{1,2}$ \qquad Yanghong Zhou$^1$ \qquad~P.~Y.~Mok$^{1,2,}$\thanks{Corresponding author}\\
$^1$The Hong Kong Polytechnic University~~$^2$Laboratory for Artificial Intelligence in Design\\
{\tt\small \{ji-hua.peng,yanghong.zhou\}@connect.polyu.hk, tracy.mok@polyu.edu.hk}
}

\begin{document}
\maketitle

\begin{abstract}
This paper presents a novel Kinematics and Trajectory Prior Knowledge-Enhanced Transformer (KTPFormer), which overcomes the weakness in existing transformer-based methods for 3D human pose estimation that the derivation of Q, K, V vectors in their self-attention mechanisms are all based on simple linear mapping. We propose two prior attention modules, namely Kinematics Prior Attention (KPA) and Trajectory Prior Attention (TPA) to take advantage of the known anatomical structure of the human body and motion trajectory information, to facilitate effective learning of global dependencies and features in the multi-head self-attention. KPA models kinematic relationships in the human body by constructing a topology of kinematics, while TPA builds a trajectory topology to learn the information of joint motion trajectory across frames. Yielding Q, K, V vectors with prior knowledge, the two modules enable KTPFormer to model both spatial and temporal correlations simultaneously. Extensive experiments on three benchmarks (Human3.6M, MPI-INF-3DHP and HumanEva) show that KTPFormer achieves superior performance in comparison to state-of-the-art methods. More importantly, our KPA and TPA modules have lightweight plug-and-play designs and can be integrated into various transformer-based networks (i.e., diffusion-based) to improve the performance with only a very small increase in the computational overhead. The code is available at: https://github.com/JihuaPeng/KTPFormer.
\end{abstract}

\section{Introduction}
\label{sec:intro}

\begin{figure}
    \centering
    \includegraphics[width=8cm]{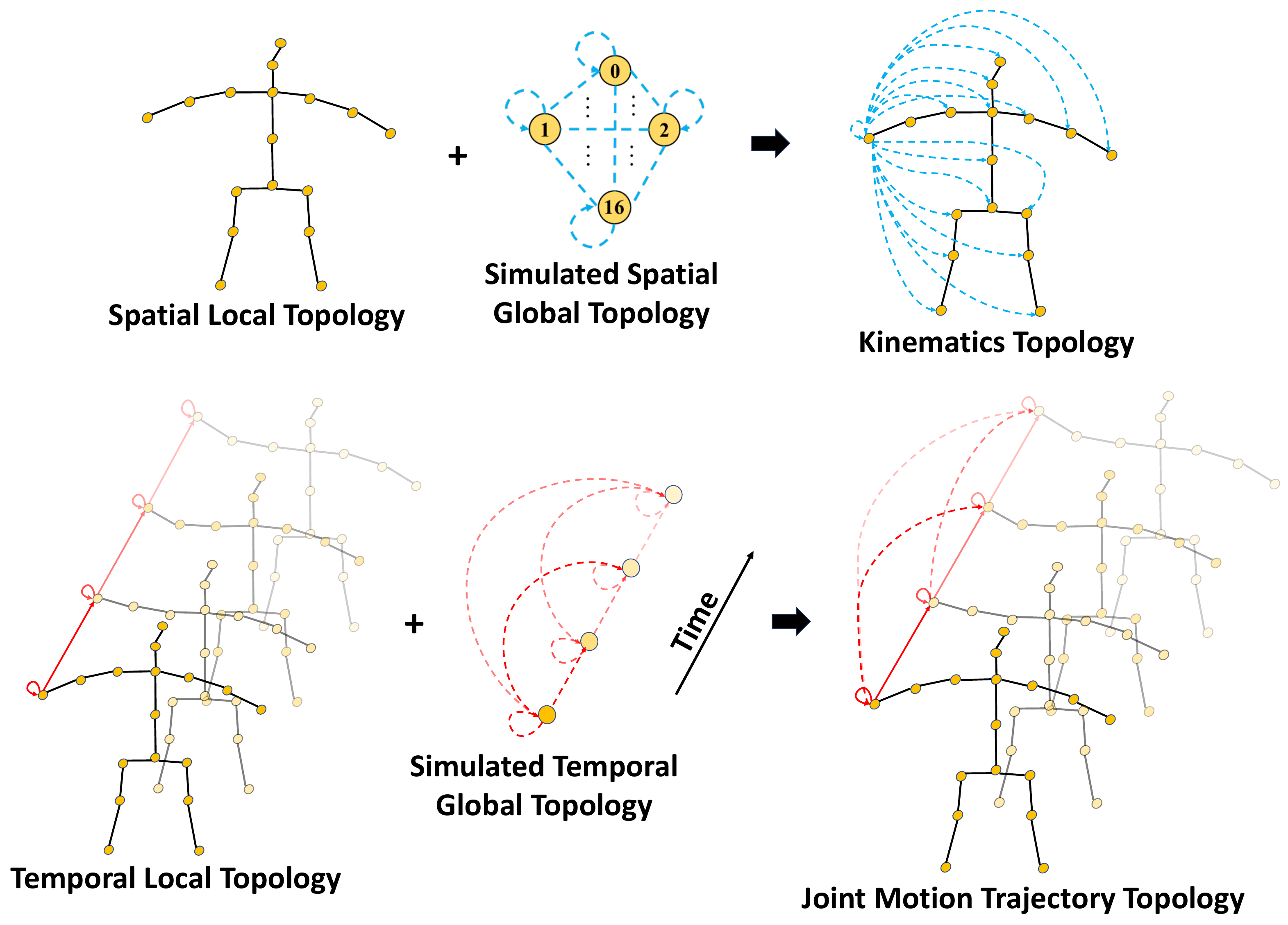}
    \vspace{-5px}
    \caption{\textbf{Top}: the spatial local topology (fixed) plus the simulated spatial global topology (learnable) to form the kinematics topology (learnable). \textbf{Bottom}: the temporal local topology (fixed) plus the simulated temporal global topology (learnable) to form the joint motion trajectory topology (learnable).}
    \vspace{-20px}
    \label{fig:spatial-temporal_graph}
\end{figure}

3D human pose estimation aims to predict the 3D coordinates of human body joints from the input of either monocular 2D images or video sequences. Because of the strong ties to action recognition \cite{liu2017enhanced,liu2018recognizing}, human-robot interaction \cite{kisacanin2005real,svenstrup2009pose} and virtual reality \cite{hagbi2010shape}, the research of 3D human pose estimation has received considerable attention in recent years. Transformer, a deep learning architecture, has revolutionized first in natural language processing (NLP) and later in other areas such as computer vision since its introduction in 2017~\cite{vaswani2017attention}. The name 'transformer' comes from the fact that these architectures use a self-attention mechanism to transform layers of inputs into layers of outputs in a way that allows the model to focus on (attend to) certain inputs. In terms of 3D pose estimation, the transformer first processes an input video into a sequence of tokens, the basic units of processing namely 2D poses, and then models the spatial-temporal relationship between tokens using \emph{multi-head self-attention} (MHSA) mechanism. 

The existing works of transformer-based methods for 3D human pose estimation \cite{zheng20213d,li2022exploiting,shan2022p,li2022mhformer,zhang2022mixste,zhao2023poseformerv2,tang20233d} mainly focus on developing novel \emph{transformer encoders}. They model either the spatial correlation between joints within each frame and the pose-to-pose or joint-to-joint temporal correlation across frames. Regardless of spatial or temporal MHSA calculation, the present transformer-based methods all use linear embedding where 2D pose sequence are tokenized into high dimensional features and treated uniformly to compute the spatial correlation between joints and the temporal correlation across frames in the spatial and temporal MHSA, respectively. This may lead to the problem of 'attention collapse', a phenomenon denoting a circumstance wherein the self-attention becomes too focused on a limited subset of input tokens while disregarding other segments of the sequence. In contrast to previous works, with the known anatomical structure of the human body as well as joint motion trajectory across frames as a priori knowledge, we propose a graph-based method to formulate such prior knowledge-attention for better learning the spatial and temporal correlations. Our graph-based prior attention mechanism is different from other existing  graph-transformer methods~\cite{zhu2021posegtac,zhao2022graformer,li2023pose,gong2023diffpose}; without modifying the transformer structure or introducing complex network, instead, we design plug-and-play modules to be placed in front of MHSA modules of a vanilla transformer. Our method is simple yet effective, highly flexible and adaptable, allowing it to be integrated into different transformer-based methods. 

To be specific, we introduce two novel prior attention modules, namely Kinematics Prior Attention (KPA) and Trajectory Prior Attention (TPA), and the key concepts are illustrated in Figure~\ref{fig:spatial-temporal_graph}. KPA first constructs a spatial \emph{local} topology based on the anatomy of the human body, as shown at the top of Figure~\ref{fig:spatial-temporal_graph}. The way these joints are physically connected to each other is fixed and is represented by solid lines. To introduce the kinematic relations among non-connected joints, we use a fully connected spatial topology to calculate the joint-to-joint attention weights, called simulated spatial global topology. In this topology, the strength of the connectivity relationship between each joint (including itself) is learnable, and thus we denote it with a dotted line in Figure~\ref{fig:spatial-temporal_graph}. We combine the spatial local topology and the simulated spatial global topology to obtain a \emph{kinematics topology}, where each joint has a learnable kinematic relationship with each other. This kinematic topological information aims to provide a priori knowledge to the spatial MHSA, enabling it to assign weights to joints based on the kinematic relationships in different actions. % By doing so, we can use the convolution to learn this kinematic topology, injecting the kinematic information into spatial tokens. 
% Also, it provides the prior knowledge for the spatial MHSA and allow it to rationally assign weights among joints based on different kinematic relationships in different actions.
Similarly, as shown in the bottom of Figure~\ref{fig:spatial-temporal_graph}, TPA connects the same joint across consecutive frames to build the temporal \emph{local} topology. Next, we construct a temporal global topology by exploiting learnable vectors (dotted line) to connect the joints among all neighboring and non-neighboring frames, which is equivalent to the computation of attention weights among all frames by self-attention, called simulated temporal global topology. Then, we combine the two topologies to obtain a new topology called \emph{joint motion trajectory topology}, which allows the network to learn both the temporal sequentiality and periodicity (joints in non-neighboring frames have similar motions to each other) for the joint motion. The temporal tokens embedded with the trajectory information will be more effectively activated in the temporal MHSA, which enhances the temporal modeling ability for MHSA. The KPA and TPA modules are combined with vanilla MHSA and MLP to form the Kinematics and Trajectory Prior Knowledge-Enhanced Transformer (KTPFormer) for 3D pose estimation, as shown in Figure~\ref{fig:architecture}.
% The experimental results show that our method can also be used as a lightweight plug-and-play module in various transformer-based networks to enhance the spatial and temporal modeling capabilities and improve the performance for 3D pose estimation.
In summary, the contributions of this paper are as follows:

\begin{itemize}
\item We propose two novel prior attention modules, KPA and TPA, which can be combined with MHSA and MLP in a simple yet effective way, forming the KTPFormer for 3D pose estimation.
\item Our KTPFormer outperforms the state-of-the-art methods on Human3.6M, MPI-INF-3DHP and HumanEva benchmarks, respectively.
\item KPA and TPA are designed as lightweight plug-and-play modules, which can be integrated into various transformer-based methods (including diffusion-based) for 3D pose estimation. Extensive experiments show that our method can effectively improve the performance without largely increasing computational resources.
% \item Our KTPFormer outperforms the state-of-the-art methods on Human3.6M, MPI-INF-3DHP and HumanEva benchmarks, respectively.
\end{itemize}

\section{Related Work}
\label{sec:formatting}

% \textbf{3D human pose estimation in monocular view.}

% 3D human pose estimation predicts the 3D coordinates of human joints from inputs of either 2D monocular images or video sequences. 

Monocular 3D human pose estimation methods can be broadly categorized into two approaches: (1) direct 3D methods \citep{li20153d,park20163d,tekin2016direct,pavlakos2018ordinal,moon2019camera,wehrbein2021probabilistic} %: these methods predict 3D joint coordinates directly from the input images. 
and (2) 2D-to-3D lifting methods \citep{martinez2017simple,hossain2018exploiting,pavllo20193d,liu2020attention,chen2021anatomy}. %: the 2D joint coordinates are first estimated in the image using 2D human pose detectors \citep{newell2016stacked,chen2018cascaded}, after which they are lifted into 3D space. All the methods reviewed below adopt the second strategy.
The second approach first estimates 2D joint coordinates from input image using 2D human pose detectors \citep{newell2016stacked,chen2018cascaded}, next they are lifted into 3D space. These methods can be further categorized as transformer-based or graph-based.

% and 2D-to-3D lifting methods. Comparing to the former, %direct approach~\citep{li20153d,park20163d,tekin2016direct,pavlakos2018ordinal,moon2019camera,wehrbein2021probabilistic}, predict 3D joint coordinates directly from the input images.
% without explicitly estimating the 2D joint coordinates.
% Most methods follow a 2D-to-3D approach~\citep{martinez2017simple,hossain2018exploiting,pavllo20193d,liu2020attention,chen2021anatomy}, in which 2D joint coordinates are first estimated and then lifted into 3D space for 3D coordinates.
% All the methods reviewed below, including transformer-based networks, graph networks and graph-transformers, adopt the 2D-to-3D approach.

% The 2D-to-3D lifting methods benefit from the robustness and accuracy of the 2D pose estimation models, thereby generally outperforming the direct 3D methods.
% \citet{martinez2017simple} utilized stacked fully connected layers with residual connections to directly lift the 2D pose into 3D space. \citet{hossain2018exploiting} predicted a sequence of 3D poses and introduce a temporal loss function to ensure temporal smoothness. \citet{pavllo20193d} proposed the temporal convolution network
% (TCN), composed of the dilated temporal convolutions, to predict the 3D pose of the central frame. 
% \citet{liu2020attention} incorporated an attention mechanism into the TCN, which identifies crucial frames and output tensors in each layer.
% \citet{chen2021anatomy} transformed the TCN into two sub-networks which predict the bone direction and bone length, respectively.

\textbf{Transformer-based networks.} Transformer was first proposed by \citet{vaswani2017attention} and showed remarkable performance in natural language processing (NLP), as the self-attention can model long-range dependencies and also capture global features. Recently, several studies on transformer-based methods for 3D human pose estimation have been reported, with Poseformer \citep{zheng20213d} being the first that predicts the 3D pose of the central frame by modeling spatial and temporal information. However, the computational burden is huge when the frame number increases. PoseformerV2 \citep{zhao2023poseformerv2} introduces a time-frequency feature to the transformer structure, efficiently extends the input sequence length, and achieves a good trade-off between speed and accuracy. % The strided transformer \citep{li2022exploiting} utilized the strided convolutions in the vanilla transformer to aggregate local information, predicting the 3D pose sequences and the central 3D pose, respectively.%
MHFormer \cite{li2022mhformer}, a transformer-based network, generates multiple hypotheses at the pose level and calculates the target 3D pose by averaging. 
% \citet{shan2022p} proposed a pre-trained model for masking the frames and joints in a 2D pose, and then loaded the pre-trained parameters onto a transformer network which predicted 3D poses. 
MixSTE \cite{zhang2022mixste} stacks spatial and temporal transformer blocks to capture spatial-temporal features alternatively and models the trajectory of joints over frame sequence. STCFormer \cite{tang20233d} slices the input joint features into two partitions and uses MHSA to encapsulate the spatial and temporal context in parallel. D3DP \cite{shan2023diffusion}, a diffusion-based method, recovers the noisy 3D poses by assembling joint-by-joint multiple hypotheses. By introducing new encoders for better modeling the spatial and temporal relations, these methods all have unavoidably changed the internal structure or altered the MHSA of the transformer, resulting in largely increased network complexity. %In this paper, a graph-based approach is proposed to enhance the spatial-temporal relation learning. 
%transformed the internal structure of the transformer, or proposed different combination ways of MHSA. 
% This research type made it difficult for transformer-based networks to achieve promising improvements in 3D pose estimation.
%Differently, we found that the global modeling capabilities of MHSA can be more effectively enhanced by introducing a prior attention mechanism to MHSA. The prior attention mechanism in the transformer networks can significantly improve the performance for 3D pose estimation, obviating the necessity for complex deformations on the transformer framework.

\textbf{Graph networks and graph-transformer methods.}
With a strong capability of capturing dynamic relations, graph neural networks are widely used in 3D human pose estimation \cite{zhao2019semantic,ci2019optimizing,cai2019exploiting,liu2020comprehensive,wang2020motion,zou2021modulated,xu2021graph,cheng2021graph,hu2021conditional,yu2023gla}. \citet{zhao2019semantic} introduced a semantic graph convolutional network (GCN) that involved a learnable mask to scale up the receptive field of convolution filters, capturing semantic information among local and global nodes. \citet{ci2019optimizing} proposed a locally connected network (LCN) to enhance the representation capability over GCN. 
% \citet{zou2021modulated} modulated the weight matrix and affinity matrix in the GCN to significantly reduce the estimation errors.
\citet{xu2021graph} developed a graph stacked hourglass network to learn different scales of human skeletal representations. \citet{yu2023gla} utilized the adaptive GCN \cite{li2018adaptive} to model global correlations and learn local joint representation by individually connected layers. 

Recently, some studies combined graph and transformer, introducing graph-transformer methods \cite{zhu2021posegtac,zhao2022graformer,li2023pose,gong2023diffpose}. PoseGTAC \cite{zhu2021posegtac} uses graph atrous convolution to learn the multi-scale information among 1-to-3 top neighbors and utilizes the graph transformer layer to capture long-range features. GraFormer \cite{zhao2022graformer} replaces the MLP in the transformer with learnable GCN layers to form the GraAttention block, which also contains MHSA. ChebGConv \cite{defferrard2016convolutional} models the implicit connection relations among non-neighboring joints. \citet{li2023pose} introduces a graph POT, where each element is the relative distance between a pair of joints, which are being encoded as the attention bias in the MHSA module. DiffPose \cite{gong2023diffpose} interlaces GCN layers \cite{zhao2019semantic} with self-attention layers as a diffusion model, which can capture spatial features between joints based on the human skeleton. Nevertheless, these graph-transformer methods \cite{zhu2021posegtac,zhao2022graformer,li2023pose,gong2023diffpose} learn merely the spatial information of individual pose, without considering temporal correlation across frames. %For example, DiffPose \cite{gong2023diffpose} also employed the TCN \cite{pavllo20193d} to extract temporal information in order to estimate 3D poses from video sequence. 
Moreover, they \cite{zhu2021posegtac,zhao2022graformer,li2023pose,gong2023diffpose} modify the structure of the transformer by introducing the graph convolution, resulting in much larger and more complex networks. %It makes these networks fail to have outstanding applicability. }

\section{Method}

\begin{figure*}
    \centering
    \includegraphics[width=15cm]{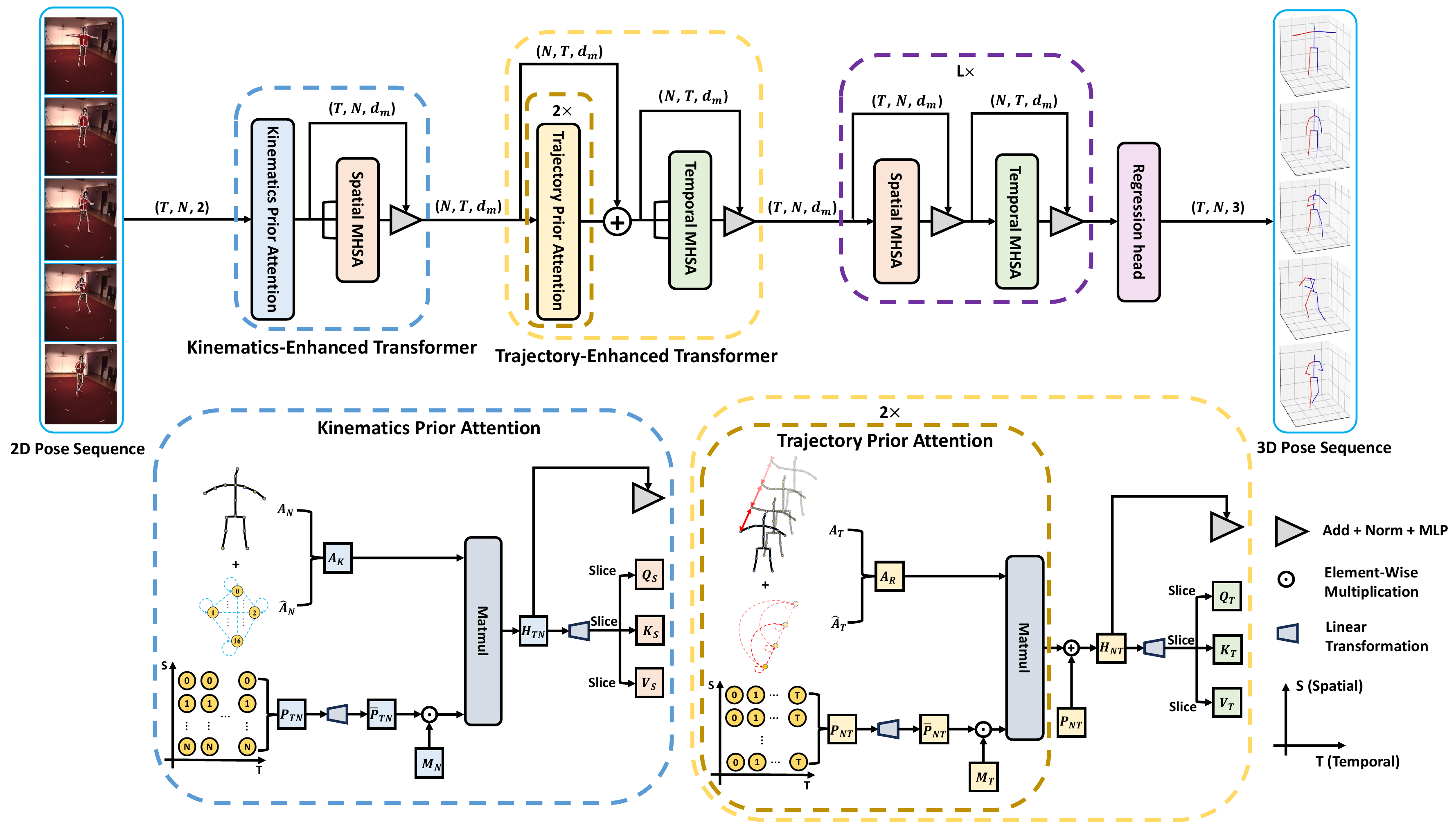}
    \vspace{-10px}
    \caption{Overview of Kinematics and Trajectory Prior Knowledge-Enhanced Transformer (KTPFormer). The input 2D pose sequence $P_{TN} \in \mathbb{R} ^{T \times N \times 2}$ with $T$ frames and $N$ joints is first fed into the Kinematics-Enhanced Transformer. KPA injects the kinematic information into the $P_{TN}$, aiming to obtain high-dimensional spatial tokens $H_{TN} \in \mathbb{R}^{T \times N \times d_{m}}$. Then, $H_{TN}$ is split into $Q_{S}$, $K_{S}$, $V_{S}$, which are then fed into the Spatial MHSA. 
    % to learn global correlation between joints. 
    The Trajectory-Enhanced Transformer takes a sequence of reshaped tokens $P_{NT} \in \mathbb{R} ^{N \times T \times d_{m}}$ as input. The stacked TPA blocks with the residual connection yield the temporal tokens $H_{NT} \in \mathbb{R}^{N \times T \times d_{m}}$, which are then sliced into $Q_{T}$, $K_{T}$, $V_{T}$ for the Temporal MHSA.}
    \vspace{-8px}
    \label{fig:architecture}
\end{figure*}

In this paper, we propose a novel Kinematics and Trajectory Prior Knowledge-Enhanced Transformer (KTPFormer), which combines kinematics and trajectory prior attentions and MHSA in a direct but effective way. Our KTPFormer can model both spatial and temporal information simultaneously. Moreover, our method preserves the inherent structure of the transformer and is more flexible. 

Our KTPFormer utilizes the seq2seq pipeline for 3D human pose estimation, which can simultaneously predict 3D pose sequence corresponding to the input 2D keypoint sequence. As shown in Figure~\ref{fig:architecture}, an input 2D pose sequence $P_{TN} \in \mathbb{R} ^{T \times N \times 2}$ is first fed into the Kinematics-Enhanced Transformer, with $T$ denotes the number of frames, $N$ denotes the number of joints, and 2 is the channel size. KPA injects the kinematic topological information into the 2D pose $P_{N} \in \mathbb{R} ^{N \times 2}$ in each frame, aiming to obtain high-dimensional spatial tokens $H_{TN} \in \mathbb{R}^{T \times N \times d_{m}}$. Next, the spatial MHSA transforms $H_{TN}$ into matrics $Q_{S}$, $K_{S}$, $V_{S}$ for learning the global correlation between joints. 
The Trajectory-Enhanced Transformer takes a sequence of reshaped tokens $P_{NT} \in \mathbb{R} ^{N \times T \times d_{m}}$ as inputs. We stack two TPA blocks with the residual connection to generate the temporal tokens $H_{NT} \in \mathbb{R}^{N \times T \times d_{m}}$ with incorporated prior information on joint motion trajectories. Next, the temporal MHSA transforms $H_{NT}$ into $Q_{T}$, $K_{T}$, $V_{T}$ for modeling global coherence among frames. The output features from Temporal MHSA are reshaped and fed into stacked spatio-temporal transformers for encoding. Finally, the regression head predicts the coordinates of the 3D pose sequence based on the learned features.

\subsection{Kinematics-Enhanced Transformer}
Kinematics-Enhanced Transformer receives the input 2D keypoint sequence and transforms them into high-dimensional spatial tokens. The 2D keypoint sequence first goes through the KPA for embedding the prior knowledge of kinematics, which is then fed into the spatial MHSA for global correlation learning between joints. 

To be specific, given a 2D pose sequence as $P_{TN} \in \mathbb{R} ^{T \times N \times 2}$, we regard each joint $p_i \in \mathbb{R}^{2}$ of a 2D pose $P_N \in \mathbb{R}^{N \times 2}$ as a keypoint patch. Next, we define a learnable transformation matrix $W \in \mathbb{R}^{2 \times d_{m}}$ to map all keypoint patches $P_{TN}$ into high-dimensional tokens $\bar{P}_{TN} \in \mathbb{R}^{T \times N \times d_{m}}$. In order to inject the prior information of kinematics into $\bar{P}_{TN}$, KPA first constructs a symmetric affinity matrix $A_{N} \in \mathbb{R}^{N \times N}$ based on the skeletal structure of the human body, namely spatial local topology, as shown in Figure~\ref{fig:spatial-temporal_graph}. If two joints are physically connected in the human body structure, the corresponding element in the affinity matrix $A_{N} \in \mathbb{R}^{N \times N}$ is non-zero and 0 otherwise. The affinity matrix $A_{N}$ can allow each 2D keypoint to learn anatomical structure information of the human body. Besides, KPA also considers the implicit kinematic relationships among adjacent and non-adjacent keypoints. Similar to the self-attention in MHSA, we establish a fully connected spatial topology, called simulated spatial global topology, as shown in Figure~\ref{fig:spatial-temporal_graph}. In this topology, all the joints are interconnected by dotted lines, indicating that the connectivity relationship between each joint is learnable. The simulated spatial global topology is denoted as an affinity matrix $\hat{A}_{N} \in \mathbb{R}^{N \times N}$, where each element is learnable. Lastly, we combine the spatial local topology $A_{N}$ with the simulated spatial global topology $\hat{A}_{N}$ to derive a kinematics topology $A_{K}$, which is shown as:
\begin{equation}
A_{K}=\frac{(A_{N}+\hat{A}_{N})+(A_{N}+\hat{A}_{N})^{\prime}}{2}
\end{equation}
where $^{\prime}$ denotes the matrix transpose, $A_{K} \in \mathbb{R}^{N \times N}$ is a learnable affinity matrix. The reason why we construct the $A_{K}$ with the above formula is that the original spatial local topology matrix $A_{N}$ is also symmetric.
% The kinematics topology is constructed based on the anatomy of human body and allows each keypoint to learn reasonable kinematic knowledge.
In order to ensure that different keypoints can learn different kinematic knowledge, we introduce a learnable weight matrix $M_{N} \in \mathbb{R}^{N \times d_m}$ and multiply it with tokens $\bar{P}_{TN} \in \mathbb{R}^{T \times N \times d_{m}}$ by element-wise multiplication, which is an economic and effective way. Thus, we can obtain the tokens $H_{TN} \in \mathbb{R}^{T \times N \times d_{m}}$ including the prior knowledge of kinematics. The formula is represented as:
\begin{equation}
H_{TN}=(M_{N} \odot \bar{P}_{TN}) A_{K}
\end{equation}
where $\odot$ represents element-wise multiplication. 
% We apply the batch normalization and activation function ReLU to $H_{TN}$ for normalizing the data and introducing non-linear features.
Moreover, we add the learnable spatial positional embedding to $H_{TN}$. After that, $H_{TN} \in \mathbb{R}^{T \times N \times d_{m}}$ is transformed into queries $Q_{S} \in \mathbb{R}^{T \times N \times d_{m}}$, keys $K_{S} \in \mathbb{R}^{T \times N \times d_{m}}$ and values $V_{S} \in \mathbb{R}^{T \times N \times d_{m}}$ by a linear transformation matrix. Then, we design a spatial MHSA ($MHSA_{S}$) to model global spatial correlation between keypoints within an identical frame. Each attention head ($i=1,...,h$) can be represented as:
\begin{equation}
head_{i}=Softmax(\frac{Q_{S}^{i}(K_{S}^{i})^{\prime}}{\sqrt{d}})V_{S}^{i}
\end{equation}
where $^{\prime}$ indicates the matrix transpose. All the attention heads $h$ are concatenated together to form the $MHSA_{S}$:
\begin{equation}
MHSA_{S}(Q_{S}, K_{S}, V_{S})=\textbf{cat}(head_1,...,head_h)W_{S}
\end{equation}
where $W_{S} \in \mathbb{R}^{d_m \times d_m}$ is the linear transformation matrix.
Concurrently, $H_{TN}$ as a residual adds the output of $MHSA_{S}$ to form the new output $H_{S} \in \mathbb{R}^{T \times N \times d_{m}}$, which is then fed into the layer normalization (LN) and MLP followed by a residual connection and LN. The formula can be represented as:
\begin{equation}
H_{S}=MHSA_{S}(Q_{S}, K_{S}, V_{S})+H_{TN}
\end{equation}
\begin{equation}
P_{NT}=MLP(LN(H_{S}))+H_{S}
\end{equation}
where $P_{NT} \in \mathbb{R}^{N \times T \times d_{m}}$ is the output of the Kinematics-Enhanced Transformer after being reshaped.

% Different from the traditional MHSA, our KPA injects the  prior kinematic knowledge into $Q_{S}$, $K_{S}$ and $V_{S}$, enhancing the global attention between joints on spatial domain.

\subsection{Trajectory-Enhanced Transformer}
Trajectory-Enhanced Transformer aims to integrate the prior trajectory information of joint motion across frames into a sequence of tokens $P_{NT} \in \mathbb{R}^{N \times T \times d_{m}}$, in which each joint is regarded as an individual token in the time dimension. TPA first connects the identical keypoints (including itself) across neighboring frames to construct the temporal local topology, as shown in Figure~\ref{fig:spatial-temporal_graph}, which is denoted as the symmetric affinity matrix $A_{T} \in \mathbb{R} ^{T \times T}$. In order to enhance the global attention of temporal coherence in the MHSA, we simulate a temporal global topology that considers the implicit temporal correlation among neighboring and non-neighboring frames. These keypoints belonging to the identical trajectory among neighboring and non-neighboring frames are connected by the learnable vector (dotted line) to form the simulated temporal global topology, as shown in Figure~\ref{fig:spatial-temporal_graph}. This topology can be expressed in the form of a learnable matrix $\hat{A}_{T} \in \mathbb{R}^{T \times T}$. Thus, the equation of joint motion trajectory topology can be represented as:
\begin{equation}
A_{R}=\frac{(A_{T}+\hat{A}_{T})+(A_{T}+\hat{A}_{T})^{\prime}}{2}
\end{equation}
where $^{\prime}$ denotes the matrix transpose, $A_{R} \in \mathbb{R}^{T \times T}$ is a learnable affinity matrix. Then, we transform $P_{NT}$ to embeddings $\bar{P}_{NT} \in \mathbb{R}^{N \times T \times d_{m}}$ by the linear transformation. Also, we utilize a learnable weight matrix $M_{T} \in \mathbb{R}^{T \times d_m}$ to allow different keypoints for different prior knowledge learning. The formula of one TPA is represented as:
\begin{equation}
TPA(P_{NT}):=(M_{T} \odot \bar{P}_{NT})A_{R}
\end{equation}
We stack two TPA blocks with a residual connection to obtain the temporal tokens $H_{NT} \in \mathbb{R}^{N \times T \times d_{m}}$ as follows:
\begin{equation}
H_{NT} = TPA(TPA(P_{NT}))+P_{NT}
\end{equation}
The learnable temporal positional embedding is then added to $H_{NT}$. After that, the $H_{NT} \in \mathbb{R}^{N \times T \times d_{m}}$ is converted into queries $Q_{T} \in \mathbb{R}^{N \times T \times d_{m}}$, keys $K_{T} \in \mathbb{R}^{N \times T \times d_{m}}$ and values $V_{T} \in \mathbb{R}^{N \times T \times d_{m}}$ by the linear transformation. We use a temporal MHSA ($MHSA_{T}$) to model the global temporal correlation between joints across all frames as follows:
\begin{equation}
head_{i}=Softmax(\frac{Q_{T}^{i}(K_{T}^{i})^{\prime}}{\sqrt{d}})V_{T}^{i}
\end{equation}

\begin{equation}
MHSA_{T}(Q_{T}, K_{T}, V_{T})=\textbf{cat}(head_1,...,head_h)W_{T}
\end{equation}
where $W_{T} \in \mathbb{R}^{d_m \times d_m}$ is the linear transformation matrix. Similar to $MHSA_{S}$, we can obtain the final output $H_{ST}$:
\begin{equation}
H_{T}=MHSA_{T}(Q_{T}, K_{T}, V_{T})+H_{NT}
\end{equation}
\begin{equation}
H_{ST}=MLP(LN(H_{T}))+H_{T}
\end{equation}
where $H_{ST} \in \mathbb{R}^{N \times T \times d_{m}}$ is the output of Trajectory-Enhanced Transformer.

\subsection{Stacked Spatio-Temporal Encoders}

After being reshaped, $H_{ST}$ is fed into the stacked spatio-temporal encoders which consist of alternating spatial and temporal transformers. The number of stacks is $L$. The sequential features are reshaped according to the type of the MHSA before fed into the encoder (spatial or temporal).

\subsection{Regression Head}
We utilize the linear layer as a regression head to predict the 3D pose sequence $\Hat{P}_{3D} \in \mathbb{R} ^{T \times N \times 3}$. The overall loss function for our network is given as:
\begin{equation}
\mathcal{L}=\mathcal{L}_{W}+\lambda_{T}\mathcal{L}_{T}+\lambda_{M}\mathcal{L}_{M}
\end{equation}
where $\mathcal{L}_{W}$ denotes the weighted mean per-joint position error (WMPJPE) loss \citep{zhang2022mixste}, $\mathcal{L}_{T}$ is the temporal consistency loss \citep{hossain2018exploiting}, and $\mathcal{L}_{M}$ indicates the mean per joint velocity error (MPJVE) loss \citep{pavllo20193d}. Here $\lambda_{T}$ and $\lambda_{M}$ are hyper-parameters.

\section{Experiments}

% We evaluate our method on three public datasets: Human3.6M \citep{ionescu2013human3}, HumanEva \citep{sigal2010humaneva} 
% and MPI-INF-3DHP \citep{mehta2017monocular}.

\subsection{Datasets and Protocols}

\textbf{Datasets.} We evaluated our model on three public datasets, namely Human3.6M \citep{ionescu2013human3}, MPI-INF-3DHP \citep{mehta2017monocular} and HumanEva \citep{sigal2010humaneva}. Human3.6M is an indoor scenes dataset with 3.6 million video frames. It has 11 professional actors, performing 15 actions under 4 synchronized camera views. Following previous work \citep{zhang2022mixste,tang20233d}, we used subjects 1, 5, 6, 7 and 8 for training, and subjects 9 and 11 for testing. MPI-INF-3DHP is also a public large-scale dataset.
% We train our model on this dataset to verify the generalization capabilities of our method. 
Following the setting of \citep{zhang2022mixste,tang20233d}, we used the area under the curve (AUC), percentage of correct keypoints (PCK) and MPJPE as evaluation metrics. HumanEva is a smaller dataset. To have a fair comparison with \citep{zheng20213d,zhang2022mixste}, we evaluated our method for actions (Walk and Jog) of subjects S1, S2, S3. 

\textbf{Protocol.} Protocol\#1 is denoted as the mean per-joint position error (MPJPE), which is the average euclidean distance in millimeters (mm) between the predicted and the ground-truth 3D joint coordinates. Protocol\#2 refers to the reconstruction error after the predicted 3D pose is aligned to the ground-truth 3D pose using procrustes analysis \citep{gower1975generalized}, denoted as P-MPJPE (mm).

\begin{table*}[t]
\caption{Quantitative comparison results with the state-of-the-art methods on Human3.6M. The 2D poses obtained by CPN \citep{chen2018cascaded} are used as inputs. \textbf{Top table}: evaluation results of MPJPE (mm); \textbf{Bottom table}: evaluation results of P-MPJPE (mm);} 
% \textbf{Bottom table}: evaluation results of MPJVE. 
%T is the number of input frames. (†) denotes using temporal information, and (*) indicates the diffusion-based methods. \textcolor{red}{Red}: Best results. \textcolor{blue}{Blue}: Runner-up results.}
\vspace{-5px}
\centering
\scalebox{0.62}{
\begin{tabular}{lc|ccccccccccccccc|c}
\hline
\textbf{MPJPE} (CPN) & \textbf{Publication} & Dir.          & Disc.         & Eat           & Greet         & Phone         & Photo         & Pose          & Pur.          & Sit           & SitD.         & Smoke         & Wait          & WalkD.        & Walk          & WalkT.        & Avg            \\ 
\hline
% TCN \cite{pavllo20193d} ($T$=243)† &CVPR'19                & 45.2          & 46.7          & 43.3          & 45.6          & 48.1          & 55.1          & 44.6          & 44.3          & 57.3          & 65.8          & 47.1          & 44.0          & 49.0          & 32.8          & 33.9          & 46.8           \\
% \citet{liu2020attention} ($T$=243)†  &CVPR'20                & 41.8          & 44.8          & 41.1          & 44.9          & 47.4          & 54.1          & 43.4          & 42.2          & 56.2          & 63.6          & 45.3          & 43.5          & 45.3          & 31.3          & 32.2          & 45.1           \\
% GLA-GCN \cite{yu2023gla} ($T$=243)† &ICCV'23                  & 41.3          & 44.3          & 40.8          & 41.8          & 45.9          & 54.1          & 42.1          & 41.5          & 57.8          & 62.9          & 45.0          & 42.8          & 45.9          & 29.4          & 29.9          & 44.4           \\
UGCN \cite{wang2020motion} ($T$=96)†  &ECCV'20       & 40.2          & 42.5          & 42.6          & 41.1          & 46.7          & 56.7          & 41.4          & 42.3          & 56.2          & 60.4          & 46.3          & 42.2          & 46.2          & 31.7          & 31.0          & 44.5           \\
% Anatomy3D \cite{chen2021anatomy} ($T$=243)† &TCSVT'21                  & 41.4          & 43.5          & 40.1          & 42.9          & 46.6          & 51.9          & 41.7          & 42.3          & 53.9          & 60.2          & 45.4          & 41.7          & 46.0          & 31.5          & 32.7          & 44.1           \\
% \citet{xu2021graph}      &CVPR'21               & 45.2          & 49.9          & 47.5          & 50.9          & 54.9          & 66.1          & 48.5          & 46.3          & 59.7          & 71.5          & 51.4          & 48.6          & 53.9          & 39.9          & 44.1          & 51.9           \\
PoseFormer \cite{zheng20213d} ($T$=81)† &ICCV'21                   & 41.5          & 44.8          & 39.8          & 42.5          & 46.5          & 51.6          & 42.1          & 42.0          & 53.3          & 60.7          & 45.5          & 43.3          & 46.1          & 31.8          & 32.2          & 44.3           \\
StridedFormer \cite{li2022exploiting} ($T$=351)† &TMM'22                  & 40.3          & 43.3          & 40.2          & 42.3          & 45.6          & 52.3          & 41.8          & 40.5          & 55.9          & 60.6          & 44.2          & 43.0          & 44.2          & 30.0          & 30.2          & 43.7           \\
GraFormer \cite{zhao2022graformer} & CVPR'22                & 45.2          & 50.8          & 48.0          & 50.0          & 54.9          & 65.0          & 48.2          & 47.1          & 60.2          & 70.0          & 51.6          & 48.7          & 54.1          & 39.7          & 43.1          & 51.8         \\
MHFormer \cite{li2022mhformer} ($T$=351)† & CVPR'22                & 39.2          & 43.1          & 40.1          & 40.9          & 44.9          & 51.2          & 40.6          & 41.3          & 53.5          & 60.3          & 43.7          & 41.1          & 43.8          & 29.8          & 30.6          & 43.0           \\
P-STMO \cite{shan2022p} ($T$=243)† &ECCV'22                 & 38.9          & 42.7          & 40.4          & 41.1          & 45.6          & 49.7          & 40.9          & 39.9          & 55.5          & 59.4          & 44.9          & 42.2          & 42.7          & 29.4          & 29.4          & 42.8           \\
MixSTE \cite{zhang2022mixste} ($T$=81)† & CVPR'22                  & 39.8          & 43.0          & 38.6          & 40.1          & 43.4          & 50.6          & 40.6          & 41.4          & 52.2          & 56.7          & 43.8          & 40.8          & 43.9          & 29.4          & 30.3          & 42.4           \\
MixSTE \cite{zhang2022mixste} ($T$=243)† & CVPR'22                  & 37.6          & 40.9          & 37.3          & 39.7          & 42.3          & 49.9          & 40.1          & 39.8          & 51.7          & 55.0          & 42.1          & 39.8          & 41.0          & 27.9          & 27.9          & 40.9           \\
DUE \cite{zhang2022uncertainty} ($T$=300)† &MM’22                 & 37.9          & 41.9          & 36.8          & 39.5          & 40.8          & 49.2          & 40.1          & 40.7          & 47.9          & 53.3          & 40.2          & 41.1          & 40.3          & 30.8          & 28.6          & 40.6           \\
GLA-GCN \cite{yu2023gla} ($T$=243)† &ICCV'23                  & 41.3          & 44.3          & 40.8          & 41.8          & 45.9          & 54.1          & 42.1          & 41.5          & 57.8          & 62.9          & 45.0          & 42.8          & 45.9          & 29.4          & 29.9          & 44.4           \\
POT \cite{li2023pose} &AAAI’23                 & 47.9           & 50.0          & 47.1           & 51.3          & 51.2          & 59.5          & 48.7          & 46.9          & 56.0          & 61.9          & 51.1          & 48.9          & 54.3          & 40.0          & 42.9          & 50.5          \\
STCFormer \cite{tang20233d} ($T$=81)†  &CVPR'23                & 40.6          & 43.0          & 38.3          & 40.2          & 43.5          & 52.6          & 40.3          & 40.1          & 51.8          & 57.7         & 42.8          & 39.8          & 42.3          & 28.0          & 29.5          & 42.0           \\
STCFormer \cite{tang20233d} ($T$=243)†  &CVPR'23                & 38.4          & 41.2          & 36.8          & 38.0          & 42.7          & 50.5          & 38.7          & 38.2          & 52.5          & 56.8          & 41.8          & 38.4          & 40.2          & 26.2          & 27.7          & 40.5           \\
DiffPose \cite{gong2023diffpose} ($T$=243)†*  &CVPR'23             & 33.2          & 36.6          & 33.0          & 35.6          & 37.6          & 45.1          & 35.7          & 35.5          & 46.4          & 49.9          & 37.3          & 35.6          & 36.5          & 24.4          & \textcolor{blue}{24.1}          & 36.9           \\
D3DP \cite{shan2023diffusion} ($T$=243)†* &ICCV'23              & \textcolor{blue}{33.0}          & \textcolor{blue}{34.8}          & \textcolor{blue}{31.7}          & \textcolor{blue}{33.1}          & \textcolor{blue}{37.5}          & \textcolor{blue}{43.7}          & \textcolor{blue}{34.8}          & \textcolor{blue}{33.6}          & \textcolor{blue}{45.7}          & \textcolor{blue}{47.8}          & \textcolor{blue}{37.0}          & \textcolor{blue}{35.0}          & \textcolor{blue}{35.0}          & \textcolor{blue}{24.3}          & \textcolor{blue}{24.1}          & \textcolor{blue}{35.4}           \\ 
\hline
Ours ($T$=81)†    &           & 39.1          & 41.9          & 37.3          & 40.1          & 44.0          & 51.3          & 39.8          & 41.0          & 51.4          & 56.0          & 43.0          & 41.0          & 42.6          & 28.8          & 29.5          & 41.8           \\
% Ours ($T$=243)†    &          & 36.8          & 39.1          & 36.7          & 37.7          & 42.2          & 48.2          & 38.4          & 39.3          & 51.5          & 56.4          & 41.4          & 39.6          & 39.6          & 27.0          & 27.4          & 40.1           \\
Ours ($T$=243)†    &          & 37.3          & 39.2          & 35.9          & 37.6          & 42.5          & 48.2          & 38.6          & 39.0          & 51.4          & 55.9          & 41.6          & 39.0         & 40.0          & 27.0          & 27.4          & 40.1           \\
Ours ($T$=243)†*   &        & \textcolor{red}{30.1} & \textcolor{red}{32.1} & \textcolor{red}{29.1} & \textcolor{red}{30.6} & \textcolor{red}{35.4} & \textcolor{red}{39.3} & \textcolor{red}{32.8} & \textcolor{red}{30.9} & \textcolor{red}{43.1} & \textcolor{red}{45.5} & \textcolor{red}{34.7} & \textcolor{red}{33.2} & \textcolor{red}{32.7} & \textcolor{red}{22.1} & \textcolor{red}{23.0} & \textcolor{red}{33.0}  \\ 
\hline\hline
\textbf{P-MPJPE} (CPN) & \textbf{Publication} & Dir.          & Disc.         & Eat           & Greet         & Phone         & Photo         & Pose          & Pur.          & Sit           & SitD.         & Smoke         & Wait          & WalkD.        & Walk          & WalkT.        & Avg            \\ 
\hline
UGCN \cite{wang2020motion} ($T$=96)† &ECCV'20                  & 31.8          & 34.3          & 35.4          & 33.5          & 35.4          & 41.7          & 31.1          & 31.6          & 44.4          & 49.0          & 36.4          & 32.2          & 35.0          & 24.9          & 23.0          & 34.5           \\
% Anatomy3D \cite{chen2021anatomy} ($T$=243)†      &TCSVT'21          & 32.6          & 35.1          & 32.8          & 35.4          & 36.3          & 40.4          & 32.4          & 32.3          & 42.7          & 49.0          & 36.8          & 32.4          & 36.0          & 24.9          & 26.5          & 35.0           \\
PoseFormer \cite{zheng20213d} ($T$=81)†   &ICCV'21                & 34.1          & 36.1          & 34.4          & 37.2          & 36.4          & 42.2          & 34.4          & 33.6          & 45.0          & 52.5          & 37.4          & 33.8          & 37.8          & 25.6          & 27.3          & 36.5           \\
StridedFormer \cite{li2022exploiting} ($T$=351)†   &TMM'22               & 32.7          & 35.5          & 32.5          & 35.4          & 35.9          & 41.6          & 33.0          & 31.9          & 45.1          & 50.1          & 36.3          & 33.5          & 35.1          & 23.9          & 25.0          & 35.2           \\
P-STMO \cite{shan2022p} ($T$=243)†  &ECCV'22                & 31.3          & 35.2          & 32.9          & 33.9          & 35.4          & 39.3          & 32.5          & 31.5          & 44.6          & 48.2          & 36.3          & 32.9          & 34.4          & 23.8          & 23.9          & 34.4           \\
MixSTE \cite{zhang2022mixste} ($T$=81)† &CVPR'22                   & 32.0          & 34.2          & 31.7          & 33.7          & 34.4          & 39.2          & 32.0          & 31.8          & 42.9          & 46.9          & 35.5          & 32.0          & 34.4          & 23.6          & 25.2          & 33.9           \\
MixSTE \cite{zhang2022mixste} ($T$=243)†  &CVPR'22                & 30.8          & 33.1          & 30.3          & 31.8          & 33.1          & 39.1          & 31.1          & 30.5          & 42.5          & 44.5          & 34.0          & 30.8          & 32.7          & 22.1          & 22.9          & 32.6           \\
DUE \cite{zhang2022uncertainty} ($T$=300)† &MM'22                 & 30.3          & 34.6          & 29.6          & 31.7          & 31.6          & 38.9          & 31.8          & 31.9          & 39.2          & 42.8          & 32.1          & 32.6          & 31.4          & 25.1          & 23.8          & 32.5           \\
GLA-GCN \cite{yu2023gla} ($T$=243)† &ICCV'23                  & 32.4          & 35.3          & 32.6          & 34.2          & 35.0          & 42.1          & 32.1          & 31.9          & 45.5          & 49.5          & 36.1          & 32.4          & 35.6          & 23.5          & 24.7          & 34.8           \\
STCFormer \cite{tang20233d} ($T$=81)†  &CVPR'23                & 30.4         & 33.8           & 31.1          & 31.7          & 33.5          & 39.5          & 30.8           & 30.0          & 41.8          & 45.8         & 34.3          & 30.1          & 32.8          & 21.9          & 23.4          & 32.7          \\
STCFormer \cite{tang20233d} ($T$=243)†  &CVPR'23                & 29.3        & 33.0           & 30.7          & 30.6          & 32.7          & 38.2          & 29.7           & 28.8         & 42.2          & 45.0         & 33.3          & 29.4          & 31.5          & 20.9          & 22.3          & 31.8          \\
D3DP \cite{shan2023diffusion} ($T$=243)†* &ICCV'23              & \textcolor{blue}{27.5}          & \textcolor{blue}{29.4}          & \textcolor{blue}{26.6}          & \textcolor{blue}{27.7}          & \textcolor{blue}{29.2}          & \textcolor{blue}{34.3}          & \textcolor{blue}{27.5}          & \textcolor{blue}{26.2}          & \textcolor{blue}{37.3}          & \textcolor{blue}{39.0}          & \textcolor{blue}{30.3}          & \textcolor{blue}{27.7}          & \textcolor{blue}{28.2}          & \textcolor{blue}{19.6}          & \textcolor{blue}{20.3}          & \textcolor{blue}{28.7}           \\ 
\hline
Ours ($T$=81)†     &          & 30.6          & 33.4          & 30.1          & 31.9          & 33.7          & 38.2          & 30.6          & 30.7          & 40.9  &44.8          & 34.4          & 30.5          & 32.7          & 22.3          & 24.0          & 32.6           \\
% Ours ($T$=243)†    &       & 30.1   &32.3 &29.8  &31.1    &32.4     &37.4   &30.1 &30.1   &41.3          &45.3          &33.4    &30.4      &31.4    &21.5  &22.4 &32.0  \\ 
Ours ($T$=243)†    &       & 30.1   & 32.3  & 29.6  & 30.8   & 32.3    & 37.3  & 30.0  & 30.2   & 41.0         & 45.3         & 33.6   & 29.9     & 31.4   & 21.5 & 22.6  & 31.9  \\ 
Ours ($T$=243)†*    &       & \textcolor{red}{24.1}   & \textcolor{red}{26.7}   & \textcolor{red}{24.2}  & \textcolor{red}{24.9}   & \textcolor{red}{27.3}    & \textcolor{red}{30.6}  & \textcolor{red}{25.2}   & \textcolor{red}{23.4}  & \textcolor{red}{34.1}         & \textcolor{red}{35.9}         & \textcolor{red}{28.1}   & \textcolor{red}{25.3}     & \textcolor{red}{25.9}   & \textcolor{red}{17.8}  & \textcolor{red}{18.8}  & \textcolor{red}{26.2} \\ 
\hline
\multicolumn{18}{l}{\footnotesize{$T$ is the number of input frames. (†) denotes using temporal information, and (*) indicates the diffusion-based methods. \textcolor{red}{Red}: Best results. \textcolor{blue}{Blue}: Runner-up results.}}
\end{tabular}}
\label{Tab:CPN}
\vspace{-10px}
\end{table*}

%\subsection{Implementation Details}
\textbf{Implementation Details.}
We implemented our method in the Pytorch framework on one GeForce RTX 3090 GPU. The input 2D keypoints were detected by 2D pose detector \citep{chen2018cascaded} or 2D ground truth.
% For a fair comparison, we set the certain input frame length T as: Human3.6M ($T$=81,243), HumanEva ($T$=27,43), MPI-INF-3DHP ($T$=27,81).
The $W$ in WMPJPE follows the setting of MixSTE \cite{zhang2022mixste}. We set the number of stacked spatio-temporal encoders $L$ to 7. Thus, the encoders contain 14 spatial and temporal transformer layers \cite{vaswani2017attention} with number of heads $h$=8, feature size $C$=512. During the training stage, we use the Adam optimizer to train our model with a batch size of 7.
% The batch size, dropout rate, and activation function are set to 7, 0.1, and GELU.
The learning rate is initialized to 0.00007 and decayed by 0.99 per epoch. 
Recently, diffusion models have been introduced in 3D pose estimation~\cite{gong2023diffpose,shan2023diffusion} and have achieved significant improvements in performance, as the diffusion process can be viewed as an augmentation method for pose data. To demonstrate the adaptability of our method, we introduced a diffusion process to our network, following the setting of D3DP~\cite{shan2023diffusion}, which also uses the transformer-based network as the backbone. We used KTPFormer as the denoiser in the D3DP \cite{shan2023diffusion}. For the design of the remaining diffusion process, our experimental parameters were set to be the same as D3DP \cite{shan2023diffusion}.
% Specifically, the input 2D keypoints and noisy 3D poses are fed into our two KPA modules to obtain two high-dimensional tokens, respectively. 
% Then, the two tokens are concatenated as inputs to generate $Q_{S}, K_{S}, V_{S}$ for the Spatial MHSA. 
% For the design of the remaining diffusion process, our experimental parameters are set to be the same as D3DP \cite{shan2023diffusion}.
% During the training and inference stages, we set the number of hypotheses, iterations and timesteps as the same as D3DP \cite{shan2023diffusion} for the fair comparison. 
% In short, we use our KTPFormer as the backbone of the denoiser in the D3DP 

\begin{table*}[t]
\centering
\caption{Quantitative comparison results of MPJPE (mm) with the state-of-the-art methods on Human3.6M using ground-truth (GT) 2D poses as inputs.}% $T$ is the number of input frames. (†) denotes using temporal information, and (*) indicates the diffusion-based methods. \textcolor{red}{Red}: Best results. \textcolor{blue}{Blue}: Runner-up results.}
\label{Tab: Comparisons on human3.6 using grountruth}
\vspace{-5px}
\scalebox{0.65}{
\begin{tabular}{lc|ccccccccccccccc|c} 
\hline
\textbf{MPJPE} (GT) & \textbf{Publication}   & Dir.          & Disc.         & Eat           & Greet         & Phone         & Photo         & Pose          & Pur.          & Sit           & SitD.         & Smoke         & Wait          & WalkD.        & Walk          & WalkT.        & Avg            \\ 
\hline
% \citet{liu2020attention} ($T$=243)†    & CVPR'20                 & 34.5          & 37.1          & 33.6          & 34.2          & 32.9          & 37.1          & 39.6          & 35.8          & 40.7          & 41.4          & 33.0          & 33.8          & 33.0          & 26.6          & 26.9          & 34.7           \\
UGCN \cite{wang2020motion} ($T$=96)†   &ECCV'20               & 23.0          & 25.7          & 22.8          & 22.6          & 24.1          & 30.6          & 24.9          & 24.5          & 31.1          & 35.0          & 25.6          & 24.3          & 25.1          & 19.8          & 18.4          & 25.6           \\
PoseGTAC \cite{zhu2021posegtac}  &IJCAI'21
& 37.2          & 42.2          & 32.6          & 38.6          & 38.0          & 44.0          & 40.7          & 35.2          & 41.0          & 45.5          & 38.2          & 39.5          & 38.2          & 29.8          & 33.0          & 38.2     \\
PoseFormer \cite{zheng20213d} ($T$=81)†  &ICCV'21             & 30.0          & 33.6          & 29.9          & 31.0          & 30.2          & 33.3          & 34.8          & 31.4          & 37.8          & 38.6          & 31.7          & 31.5          & 29.0          & 23.3          & 23.1          & 31.3           \\
StridedFormer \cite{li2022exploiting} ($T$=351)†   &TMM'22           & 27.1          & 29.4          & 26.5          & 27.1          & 28.6          & 33.0          & 30.7          & 26.8          & 38.2          & 34.7          & 29.1          & 29.8          & 26.8          & 19.1          & 19.8          & 28.5           \\
GraFormer \cite{zhao2022graformer} & CVPR'22                & 32.0          & 38.0          & 30.4          & 34.4         & 34.7          & 43.3          & 35.2          & 31.4          & 38.0          & 46.2          & 34.2          & 35.7          & 36.1          & 27.4          & 30.6          & 35.2         \\
MHFormer \cite{li2022mhformer} ($T$=351)†  & CVPR'22           & 27.7          & 32.1          & 29.1          & 28.9          & 30.0          & 33.9          & 33.0          & 31.2          & 37.0          & 39.3          & 30.0          & 31.0          & 29.4          & 22.2          & 23.0          & 30.5           \\
P-STMO \cite{shan2022p} ($T$=243)†      &ECCV'22        & 28.5          & 30.1          & 28.6          & 27.9          & 29.8          & 33.2          & 31.3          & 27.8          & 36.0          & 37.4          & 29.7          & 29.5          & 28.1          & 21.0          & 21.0          & 29.3           \\
DUE \cite{zhang2022uncertainty} ($T$=300)† &MM’22                 & 22.1          & 23.1          & 20.1          & 22.7          & 21.3          & 24.1         & 23.6          & 21.6          & 26.3          & 24.8          & 21.7          & 21.4          & 21.8          & 16.7          & 18.7          & 22.0           \\
MixSTE \cite{zhang2022mixste} ($T$=81)†   &CVPR'22             & 25.6          & 27.8          & 24.5          & 25.7          & 24.9          & 29.9          & 28.6          & 27.4          & 29.9          & 29.0          & 26.1          & 25.0          & 25.2          & 18.7          & 19.9          & 25.9           \\
MixSTE \cite{zhang2022mixste} ($T$=243)†  &CVPR'22            & 21.6          & 22.0          & 20.4          & 21.0          & 20.8          & 24.3          & 24.7          & 21.9          & 26.9          & 24.9          & 21.2          & 21.5          & 20.8          & 14.7          & 15.7          & 21.6           \\ 
POT \cite{li2023pose} &AAAI’23                 & 32.9            & 38.3          & 28.3           & 33.8          & 34.9          & 38.7          & 37.2          & 30.7          & 34.5          & 39.7          & 33.9          & 34.7          & 34.3          & 26.1          & 28.9          & 33.8          \\
STCFormer \cite{tang20233d} ($T$=81)†  &CVPR'23                & 26.2          & 26.5          & 23.4          & 24.6          & 25.0          & 28.6          & 28.3          & 24.6          & 30.9          & 33.7         & 25.7          & 25.3          & 24.6          & 18.6          & 19.7          & 25.7           \\
STCFormer \cite{tang20233d} ($T$=243)†  &CVPR'23                & 21.4          & 22.6          & 21.0          & 21.3          & 23.8          & 26.0          & 24.2          & 20.0          & 28.9          & 28.0         & 22.3          & 21.4          & 20.1          & 14.2          & 15.0          &  22.0          \\
GLA-GCN \cite{yu2023gla} ($T$=243)†  &ICCV'23                &  20.1         & 21.2          & 20.0          & 19.6          & 21.5         & 26.7          & 23.3          & 19.8          & 27.0          & 29.4         & 20.8          & 20.1          & 19.2          & \textcolor{blue}{12.8}          & 13.8          & 21.0           \\
DiffPose \cite{gong2023diffpose} ($T$=243)†* &CVPR'23              & \textcolor{red}{18.6}         & 19.3          & \textcolor{red}{18.0}         & 18.4          & \textcolor{red}{18.3}          & 21.5          & 21.5          & 19.1          & 23.6         & 22.3          & \textcolor{blue}{18.6}          & 18.8          & 18.3          & \textcolor{blue}{12.8}          & 13.9          & 18.9           \\ 
D3DP \cite{shan2023diffusion} ($T$=243)†* &ICCV'23              & \textcolor{blue}{18.7}          & \textcolor{blue}{18.2}          & 18.4          & \textcolor{blue}{17.8}          & \textcolor{blue}{18.6}         & \textcolor{blue}{20.9}          & \textcolor{blue}{20.2}          & \textcolor{red}{17.7}          & 23.8         & \textcolor{red}{21.8}          & \textcolor{red}{18.5}          & \textcolor{blue}{17.4}          & \textcolor{blue}{17.4}          & 13.1          & \textcolor{blue}{13.6}          & \textcolor{blue}{18.4}           \\ 
\hline
Ours ($T$=81)†       &    & 22.5          & 22.4          & 21.3          & 21.4          & 21.2          & 25.5          & 24.2          & 22.4          & 24.4          & 27.5          & 22.7          & 21.4          & 21.7          & 16.3          & 17.3          & 22.2           \\
Ours ($T$=243)†      &    & 19.6   & 18.6   &18.5    & 18.1    & 18.7    & 22.1        & 20.8    & \textcolor{blue}{18.3}    & \textcolor{red}{22.8}   & 22.4   & 18.8  & 18.1 & 18.4   & 13.9 & 15.2 & 19.0  \\
Ours ($T$=243)†*    &    & 18.8    & \textcolor{red}{17.4}   & \textcolor{blue}{18.1}   & \textcolor{red}{17.7}    & \textcolor{red}{18.3}    & \textcolor{red}{20.6}       & \textcolor{red}{19.6}    & \textcolor{red}{17.7}    & \textcolor{blue}{23.3}   & \textcolor{blue}{22.0}   & 18.7  & \textcolor{red}{17.0}   & \textcolor{red}{16.8}   & \textcolor{red}{12.4}   & \textcolor{red}{13.5}  & \textcolor{red}{18.1} \\ 
\hline
\multicolumn{17}{l}{\footnotesize{$T$ is the number of input frames. (†) denotes using temporal information, and (*) indicates the diffusion-based methods. \textcolor{red}{Red}: Best results. \textcolor{blue}{Blue}: Runner-up results.}}
\end{tabular}}
\vspace{-10px}
\end{table*}

\begin{table}
\caption{Performance comparisons on MPI-INF-3DHP} %with PCK, AUC and MPJPE. }%The ↑ denotes the higher, the better, the ↓ denotes the lower, the better.}
% (*) indicates the diffusion-based methods. 
% \textcolor{red}{Red}: Best results. \textcolor{blue}{Blue}: Runner-up results.}
\centering
\vspace{-5px}
\scalebox{0.65}{
\begin{tabular}{lc|ccc} 
\hline
\textbf{Method}     & \textbf{Publication}            & PCK↑ & AUC↑ & MPJPE↓  \\ 
\hline
UGCN \cite{wang2020motion} ($T$=96)  &ECCV'20  & 86.9 & 62.1 & 68.1    \\
% Anatomy3D \cite{chen2021anatomy} ($T$=81) &TCSVT'21  & 87.9 & 54.0 & 78.8    \\
% gong2021poseaug~~      & 88.6 & 57.3 & 73.0    \\
PoseFormer \cite{zheng20213d} ($T$=9)  &ICCV'21     & 88.6 & 56.4 & 77.1    \\
MHFormer \cite{li2022mhformer} ($T$=9)  &CVPR'22  & 93.8 & 63.3 & 58.0    \\
MixSTE \cite{zhang2022mixste} ($T$=27)  &CVPR'22  & 94.4 & 66.5 & 54.9    \\
P-STMO \cite{shan2022p} ($T$=81)   &ECCV'22          & 97.9 & 75.8 & 32.2    \\
Diffpose \cite{gong2023diffpose} ($T$=81) &CVPR'23        & 98.0 & 75.9 & 29.1    \\
D3DP \cite{shan2023diffusion} ($T$=243)  &ICCV'23           & 98.0 & 79.1 & 28.1    \\
PoseFormerV2 \cite{zhao2023poseformerv2} ($T$=81)    &CVPR'23   & 97.9 & 78.8 & 27.8    \\
GLA-GCN \cite{yu2023gla} ($T$=81)    &ICCV'23       & 98.5 & 79.1 & 27.7    \\
STCFormer \cite{tang20233d} ($T$=27)    &CVPR'23     & 98.4 & 83.4 & 24.2    \\
STCFormer \cite{tang20233d} ($T$=81)    &CVPR'23     & 98.7 & 83.9 & 23.1    \\ 
\hline
Ours ($T$=27)      &       & \textcolor{red}{98.9}   & \textcolor{blue}{84.4}   & \textcolor{blue}{19.2}    \\
Ours ($T$=81)      &       & \textcolor{red}{98.9}   & \textcolor{red}{85.9}    & \textcolor{red}{16.7}    \\
\hline
\multicolumn{5}{l}{\footnotesize{Metric ↑ denotes the higher, the better, ↓ denotes the lower, the better.}}
\end{tabular}}
\label{Tab: 3DHP}
\vspace{-5px}
\end{table}

\begin{table}
\centering
\caption{The MPJPE evaluation results on HumanEva testset.} 
% \textcolor{red}{Red}: Best results. \textcolor{blue}{Blue}: Runner-up results.}
\label{Tab: evaluation results on HumanEva}
\centering
\vspace{-5px}
\scalebox{0.65}{
\begin{tabular}{l|ccc|ccc|c} 
\hline
\multicolumn{1}{c|}{\textbf{\textbf{Method}}} & \multicolumn{1}{l}{}   & Walk                   & \multicolumn{1}{l|}{}  & \multicolumn{1}{l}{}   & Jog                    & \multicolumn{1}{l|}{}  & \textbf{Avg}            \\
                                              & S1                     & S2                     & S3                     & S1                     & S2                     & S3                     & \multicolumn{1}{l}{}    \\ 
\hline
TCN \cite{pavllo20193d} ($T$=81)                                    & \textcolor{blue}{13.1} & \textcolor{red}{10.1}  & 39.8                   & \textcolor{red}{20.7}  & \textcolor{blue}{13.9} & 15.6                   & 18.9                    \\
PoseFormer \cite{zheng20213d} ($T$=43)                             & 16.3                   & \textcolor{blue}{11.0} & 47.1                   & 25.0                   & 15.2                   & \textcolor{blue}{15.1} & 21.6                    \\
MixSTE \cite{zhang2022mixste} ($T$=43)                         & 20.3                   & 22.4                   & 34.8                   & 27.3                   & 32.1                   & 34.3                   & 28.5                    \\ 
\hline
Ours ($T$=43)                                    & 16.5                   & 13.9                   & \textcolor{blue}{19.9} & 25.3                   & 15.9                   & 16.5                   & \textcolor{blue}{18.0}  \\
Ours ($T$=27)                                    & \textcolor{red}{12.3}  & 11.5                   & \textcolor{red}{19.5}  & \textcolor{blue}{20.9} & \textcolor{red}{13.1}  & \textcolor{red}{14.5}  & \textcolor{red}{15.3}   \\
\hline
\end{tabular}}
\label{Tab: HumanEva}
\vspace{-10px}
\end{table}

\begin{table}
\centering
\caption{Results of ablation study of each module in our KTPFormer on Human3.6M dataset.} 
% The input is ground-truth 2D poses.}
\label{Tab:Ablation study of each component}
\vspace{-5px}
\scalebox{0.65}{
\begin{tabular}{l|ccc} 
\hline
\textbf{Method}      & MPJPE (mm)                      &Parameters (M) & \multicolumn{1}{l}{FLOPs (M)}  \\ 
\hline
Baseline    & 21.8                           & 33.6506       & 139038                        \\
+KPA(w/o prior)                               & 21.5                  & 33.6501        & 139042                         \\
+KPA(w/o global)                          & 20.7                  & 33.6501        & 139042  \\
+KPA        & 20.0                           & 33.6501       & 139042                        \\
+TPA(w/o prior)                               & 21.4                  & 33.6527        & 139055                         \\
+TPA(w/o global)                          & 20.5                  & 33.6527        & 139055                         \\
+TPA        & 19.7                           & 33.6527       & 139055                        \\
+KPA(w/o prior)+TPA(w/o prior)~~         & 21.4                  & 33.6522        & 139059                         \\
+KPA(w/o global)+TPA(w/o global) & 20.0                  & 33.6522        & 139059                         \\
+KPA+TPA~ ~ & \textcolor{red}{19.0} & 33.6522       & 139059                        \\
% +KPA+TPA+D3DP \cite{shan2023diffusion}~ ~ & \textcolor{red}{18.1} &        &                         \\
\hline
\end{tabular}}
\vspace{-5px}
\end{table}

\begin{table}
\centering
\caption{Results of ablation study involving different combinations of KPA and TPA in the network.}
\label{Tab: Ablation study of different combinations of modules}
\vspace{-5px}
\scalebox{0.65}{
\begin{tabular}{l|ccc} 
\hline
\multicolumn{1}{l|}{\textbf{\textbf{Method}}} & MPJPE (mm)            & Parameters (M) & \multicolumn{1}{l}{FLOPs (M)}  \\ 
\hline
Baseline          & 21.8                  & 33.6506        & 139038                         \\
United Mode  (UMD)      & 20.0                  & 33.6522        & 139059                         \\
Parallel Mode (PMD)             & 19.8                  & 33.6512        & 139051                         \\
Separate Mode-S (SMD-S)              & 20.4                  & 33.6512        & 139051                         \\
Separate Mode (SMD)          & \textcolor{red}{19.0} & 33.6522        & 139059                         \\
\hline
\end{tabular}}
\vspace{-10px}
\end{table}

\subsection{Comparison with State-of-the-art Methods}
\textbf{Results based on Human3.6M.}
We compared our results with those of recent state-of-the-art methods based on the dataset Human3.6M. As shown in Table~\ref{Tab:CPN}, our method (diffusion-based) achieves the state-of-the-art (SOTA) result 33.0mm in MPJPE and 26.2mm in P-MPJPE using the 2D poses detected by CPN \citep{chen2018cascaded} as inputs. Our method (diffusion-based) outperforms D3DP \cite{shan2023diffusion} by 2.4mm under MPJPE and 2.5mm under P-MPJPE with the same settings (the number of frames, hypotheses, and iterations) as D3DP \cite{shan2023diffusion}.
% More importantly, the KTPFormer (diffusion-based) reaches the best reported performances in 15 actions under both MPJPE and P-MPJPE.
This demonstrates that our network can serve as an excellent backbone for diffusion-based methods, effectively improving the performance for 3D pose estimation. Besides, we obtain the best results 40.1mm under $T$=243 setting and 41.8mm under $T$=81 setting in MPJPE among all methods that are not diffusion-based. 

% For the result of MPJVE, our method outperforms all other methods and achieves the best results in all actions.

Table~\ref{Tab: Comparisons on human3.6 using grountruth} compares our results with those of SOTA models using ground-truth 2D poses as inputs. Our method (diffusion-based) achieves the SOTA result 18.1mm, with the same settings (the number of frames, hypotheses, and iterations) as D3DP \cite{shan2023diffusion}. On the other hand, we obtain the best result 19.0mm under $T$=243 setting and 22.2mm under $T$=81 setting in MPJPE without diffusion process. Compared to GLA-GCN \cite{yu2023gla}, there is a noticeable improvement (21.0→19.0mm) with $T$=243. Under $T$=81 setting, our method outperforms the second-best result by 3.5mm. 

% (25.7→22.2mm). 

% Table~\ref{Tab: Comparisons on human3.6 using grountruth} verifies the effectiveness of our method for different types of input.

\textbf{Results based on MPI-INF-3DHP.} We evaluated the performance on MPI-INF-3DHP dataset to verify the generalization capability of our method. Following previous work \citep{zhao2023poseformerv2,tang20233d}, we trained our model with ground-truth 2D poses as inputs.
% on this dataset. % For a fair comparison, we set the number of input frames to 27 and 81.
Table~\ref{Tab: 3DHP} shows the comparison results on the MPI-INF-3DHP test set. Our method with $T$=81 achieves the to-date best result with PCK of 98.9\%, AUC of 85.9\% and MPJPE of 16.7mm, outperforming the existing SOTA models by 0.2\% in PCK, 2.0\% in AUC and 6.4mm in MPJPE. Moreover, our method with $T$=27 also surpasses all other methods in all three metrics.
% Notably, our method ($T$=27) exceeds MixSTE \cite{zhang2022mixste} by 17.9\% in AUC and 35.7mm under MPJPE, signifying a large margin of error decrease (65\%).
These results demonstrate the strong generalization capability of our method.

\textbf{Results based on HumanEva.} Table~\ref{Tab: HumanEva} shows the performance comparison between ours and other methods on HumanEva dataset. Our method yields the best result of 15.3mm under $T$=27. Also, our method is superior to other algorithms under $T$=81. 
% Compared with MixSTE \citep{zhang2022mixste}, we achieve 36.8\% improvement (28.5→18.0mm) under $T$=81. 
Due to the short video length in HumanEva, our method gives better results under $T$=27 than $T$=81. These results highlight the effectiveness of our method on small datasets.

\subsection{Ablation Study}

\textbf{Effect of each module}
To verify the effectiveness of the proposed modules, we conducted ablation experiments on Human3.6M ($T$=243) using ground-truth 2D poses as inputs. Table~\ref{Tab:Ablation study of each component} presents the results of ablation study of each module. Our baseline network utilizes a linear layer to lift the 2D pose sequence to the high-dimensional space and then exploits the stacked spatio-temporal encoders ($L$=8) to predict the 3D pose sequence. 
% reaching 22.1mm of MPJPE. 
As shown, the incorporation of KPA and TPA brings 1.8mm and 2.1mm of MPJPE drops, respectively. The prior knowledge of KPA and TPA contribute 1.5mm and 1.7mm of error drop, respectively. The global topologies of KPA and TPA yield reduction in MPJPE of 0.7mm and 0.8mm, respectively. 
% The introduction KPA and TPA brings 2.1mm and 2.4mm of MPJPE drops, respectively.
% After the KPA replaces the linear layer, the MPJPE is improved by 2.1mm and there is a small drop in the number of parameters. Also, the TPA contributes 2.4mm of error drop, proving the efficacy of exploiting the trajectory prior knowledge. 
With both KPA and TPA modules, the performance has improved 2.8mm. More remarkably, the number of parameters and FLOPs merely increase by 0.0016M and 21M, respectively, showing that our method is both effective and efficient.

% \begin{table}
% \centering
% \caption{Ablation study of each module in our KPTFormer on Human3.6M dataset. The input is ground-truth 2D poses with 243 receptive fields.}
% \label{Tab:Ablation study of each component}
% \scalebox{0.7}{
% \begin{tabular}{l|ccc} 
% \hline
% \textbf{Method}      & MPJPE (mm)                      & Parameters (M) & \multicolumn{1}{l}{FLOPs (M)}  \\ 
% \hline
% Baseline    & 22.1                           & 33.6506       & 139038                        \\
% +KPA        & 20.0                           & 33.6501       & 139042                        \\
% +TPA        & 19.7                           & 33.6527       & 139055                        \\
% +KPA+TPA~ ~ & \textcolor{red}{19.0} & 33.6522       & 139059                        \\
% \hline
% \end{tabular}}
% \end{table}

% \begin{figure}
%     \centering
%     %\includegraphics[width=8cm]{Figures/Comparison between PEformer and others_cut.pdf}
%     \includegraphics[width=8cm]{Figure_PDF/crop_Visualizations_attention_maps_v4.pdf}
%     \vspace{-5px}
%     \caption{Comparison of visualization results and attention maps between ours and MixSTE \cite{zhang2022mixste}. The x-axis and y-axis correspond to the queries and the predicted outputs, respectively.}
%     \vspace{-5px}
%     \label{fig:visualizations}
% \end{figure}

\textbf{Effect on different combinations of KPA and TPA.} 
We analyzed the impacts to performance for four different combinations of KPA and TPA, including the United Mode (UMD), the Separate Mode (SMD), the Separate Mode-S and the Parallel Mode (PMD). UMD indicates that the output of the KPA is fed into the two TPA blocks with a residual connection, followed by stacked spatio-temporal encoders. SMD represents that KPA is followed by spatial MHSA and two TPA blocks with a residual connection are followed by temporal MHSA. The SMD-S differs from the SMD in that only one TPA block is followed by temporal MHSA. For PMD, the input is fed into TPA and Kinematics-Enhanced Transformer simultaneously, and the outputs of them are then added together and fed into temporal MHSA. 
% We evaluate the four modes on Human3.6M under $T$=243 frames. 
Table~\ref{Tab: Ablation study of different combinations of modules} shows the comparison results of Human3.6M with $T$=243 frames between the four modes.
% The SMD yields the best results using ground-truth 2D poses as inputs.
The MPJPE result of UMD is worse than that of SMD because the features from KPA are fed directly into TPA, which leads to the confusion of spatial and temporal information. The comparison between PMD and SMD illustrates that TPA is more suitable to inject the trajectory information into high-dimensional tokens, rather than the initial 2D pose sequence. Besides, KPA and TPA should be independently followed by spatial and temporal MHSA, without interfering with each other.
% without introducing other information to cause disruption. 
The comparison between the SMD-S and SMD indicates that the stacked TPA blocks can inject the prior information into pose tokens more effectively. 

\subsection{Qualitative Analysis and Discussion}

\begin{figure}
    \centering
    \includegraphics[width=7.8cm]{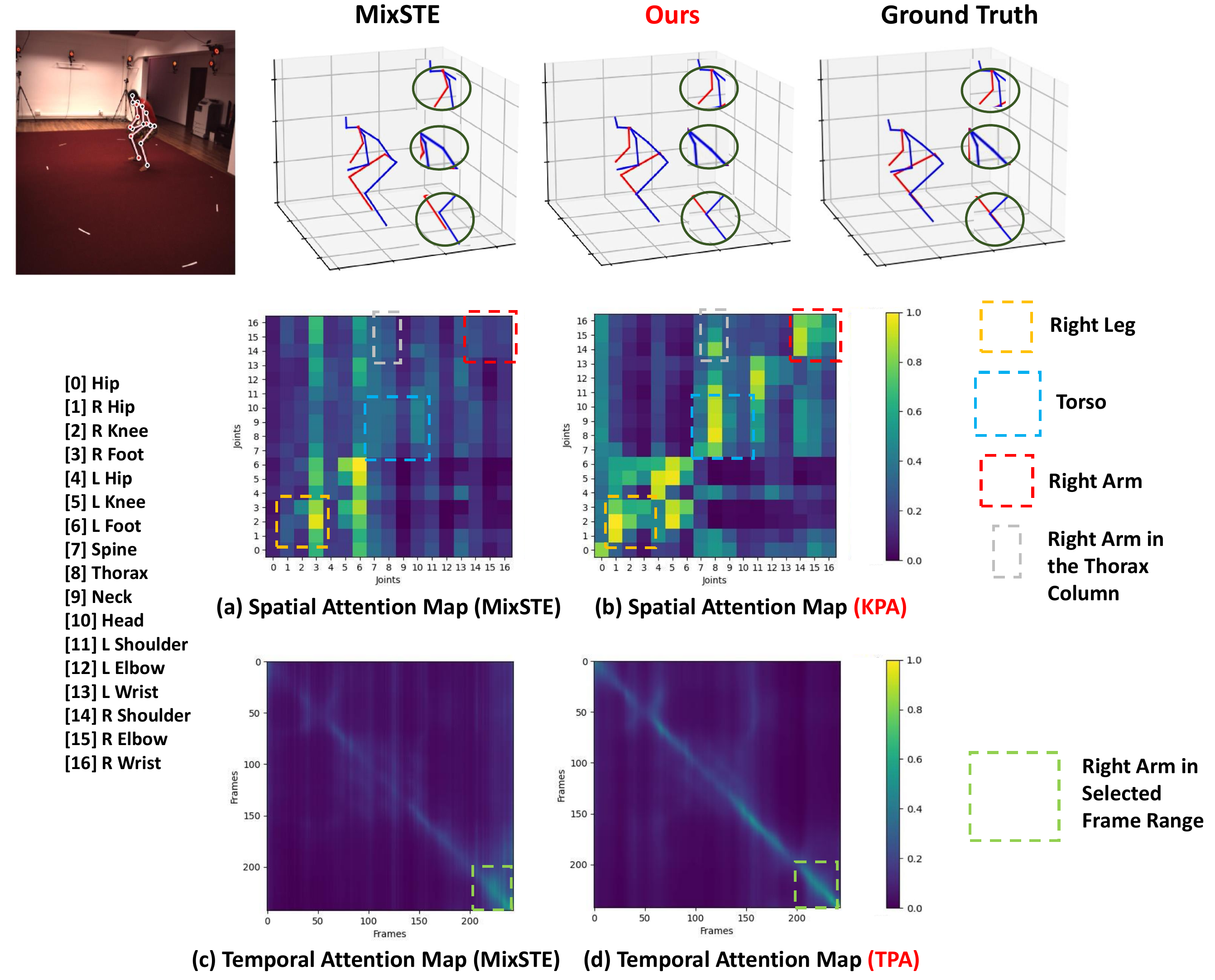}
    \vspace{-5px}
    \caption{Comparison of visualization results and attention maps between ours and MixSTE \cite{zhang2022mixste}. The x-axis and y-axis correspond to the queries and the predicted outputs, respectively.}
    \vspace{-5px}
    \label{fig:visualizations}
\end{figure}

We visualize in Figure~\ref{fig:visualizations} the 3D pose estimation results and attention maps so as to validate the efficacy of our method in comparison to MixSTE \cite{zhang2022mixste}. As shown, the spatial and temporal attention outputs from different heads are both averaged to show the distribution of attention weights of joints and frames. Figure~\ref{fig:visualizations}(a) illustrates the phenomenon of attention collapse where the attention weights become highly concentrated on the right and left foot regions with other joints ($i.e.$, torso and right arm) being ignored in the spatial attention map of MixSTE \cite{zhang2022mixste},
% unreasonable attention weight allocation to the right arm, right leg and torso in the spatial attention map of MixSTE \cite{zhang2022mixste}, 
leading to poor predicted results of 3D pose (top of Figure~\ref{fig:visualizations}). In contrast, the spatial attention weights (Figure~\ref{fig:visualizations}(b)) are activated by KPA in regions of right arm, right leg and torso. In particular, the three joints of the right arm exhibit stronger attention weights in the thorax column, owing to the anatomical connection between the right hand and the torso. The attention allocation is, therefore, more reasonable (Figure~\ref{fig:visualizations}(b)), contributing to an enhanced performance of 3D pose predicted by our method. Moreover, Figure~\ref{fig:visualizations}(c) depicts the averaged temporal attention weights of the three joints of right arm. In contrast, TPA (Figure~\ref{fig:visualizations}(d)) yields stronger temporal correlations along the \textit{diagonal} as it connects the consecutive frames and small range of non-adjacent frames. 
In cases when input video records human motion of normal speed and the video has a high frame rate (e.g., 50 fps), the joints within a small range of frames would likely exhibit very small motion variations.
% We are referring to input videos with high frame rate ($e.g.$, 50 fps) for normal-speed human motions, e.g. in above datasets, the joints within a small range of frames would likely exhibit very small motion variations.
Hence, such enhanced temporal attention can improve prediction accuracy.

% The enhanced temporal attention also contributes to the performance improvement of the right arm.

\begin{table}
\centering
\caption{Comparative results for various 3D pose estimators trained with and without KPA and TPA on Human3.6M dataset.}
% (‡): Our implementation.}
\label{Tab: Comparison results for different pose estimators}
\vspace{-5px}
\scalebox{0.63}{
\begin{tabular}{l|ccc} 
\hline
\textbf{Method}            & MPJPE (mm)                                                                       & Parameters (M)                                               & FLOPs (M)                                                          \\ 
\hline
PoseFormer \cite{zheng20213d} ($T$=81)          & 31.3                                                                             & 9.558                                                        & 815.522                                                            \\
+KPA+TPA                   & \textcolor{red}{28.8 \textsuperscript{(-2.5)}}                                   & 9.560 \textcolor[rgb]{0,0.502,0}{\textsuperscript{(+0.02)}}  & 815.885~\textcolor[rgb]{0,0.502,0}{\textsuperscript{(+0.363)}}     \\ 
\hline
StridedTransformer \cite{li2022exploiting} ($T$=351) & 28.5                                                                             & 3.979                                                        & 801.093                                                            \\
+KPA+TPA                   & \textcolor{red}{27.4~}\textcolor{red}{\textsuperscript{(-1.1)}}                  & 3.980 \textcolor[rgb]{0,0.502,0}{\textsuperscript{(+0.01)}}  & 801.859~\textcolor[rgb]{0,0.502,0}{\textsuperscript{(+0.766)}}     \\ 
\hline
MHFormer \cite{li2022mhformer} ($T$=243)           & 30.9                                                                             & 24.767                                                       & 4826.854                                                           \\
+KPA+TPA                   & \textcolor{red}{29.1 }\textcolor{red}{\textsuperscript{(-1.8)}}\textcolor{red}{} & 24.773~\textcolor[rgb]{0,0.502,0}{\textsuperscript{(+0.06)}} & 4829.873 \textcolor[rgb]{0,0.502,0}{\textsuperscript{(+3.019)}}    \\ 
\hline
MixSTE \cite{zhang2022mixste} ($T$=243)            & 21.6                                                                             & 33.650                                                       & 139038.488                                                         \\
+KPA+TPA                   & \textcolor{red}{19.0 \textsuperscript{(-2.6)}}                                   & 33.652~\textcolor[rgb]{0,0.502,0}{\textsuperscript{(+0.02)}} & 139059.638~\textcolor[rgb]{0,0.502,0}{\textsuperscript{(+21.15)}}  \\ 
\hline
STCFormer \cite{tang20233d} ($T$=81)          & 25.7                                                                             & 4.747                                                        & 6535.219                                                           \\
+KPA+TPA                   & \textcolor{red}{25.1 \textsuperscript{(-0.6)}}                                   & 4.748 \textcolor[rgb]{0,0.502,0}{\textsuperscript{(+0.01)}}  & 6541.565~\textcolor[rgb]{0,0.502,0}{\textsuperscript{(+6.346)}}    \\ 
\hline
% PoseFormerV2 \cite{zhao2023poseformerv2} ($T$=81)‡        & 37.3                                                                             & 14.351                                                       & 528.622                                                            \\
% +KPA+TPA                   & \textcolor{red}{36.3 \textsuperscript{(-1.0)}}                                   & 14.351~\textcolor[rgb]{0,0.502,0}{\textsuperscript{(+0.00)}} & 528.655~\textcolor[rgb]{0,0.502,0}{\textsuperscript{(+0.033)}}     \\
% \hline
\end{tabular}}
\vspace{-15px}
\end{table}

%\subsection{Adaptable to Different 3D Pose Estimators}
\textbf{Adaptable to Different 3D Pose Estimators.}
Our KPA and TPA are generic and can be applied in various transformer-based 3D pose estimators. To verify the adaptability, we selected five transformer-based 3D pose estimators as backbones. We removed the linear embedding before the first spatial encoder and put KPA in front of the first MHSA in these models. We used TPA to encode the features of different poses across frames on \cite{zheng20213d,li2022exploiting,li2022mhformer} and different joints across frames on \cite{zhang2022mixste,tang20233d}.
% PoseFormer \cite{zheng20213d}, StridedTransformer \cite{li2022exploiting}, MHformer \cite{li2022mhformer} and PoseFormerV2 \cite{zhao2023poseformerv2}. 
% On the other hand, we utilize the TPA to encode the feature of different joints across frames on \cite{zhang2022mixste,tang20233d}.
% MixSTE \cite{zhang2022mixste} and STCFormer \cite{tang20233d}. 
We trained these models on the Human3.6M dataset using 2D ground-truth poses as inputs. As shown in Table~\ref{Tab: Comparison results for different pose estimators}, our method brings about noticeable improvements in all the models in terms of MPJPE (mm), with very slight increases in the number of parameters and FLOPs, indicating that our KPA and TPA modules are lightweight and plug-and-play to different models for 3D pose estimation.

\section{Conclusion}
In this paper, we develop a Kinematics and Trajectory Prior Knowledge-Enhanced Transformer (KTPFormer), which explores two novel prior attention mechanisms (KPA and TPA) for 3D pose estimation.
% Specifically, Kinematic Prior Attention (KPA) 
% Specifically, KPA constructs a kinematics topology to inject the kinematics prior knowledge into spatial tokens.
% % Trajectory Prior Attention (TPA) 
% TPA incorporate the prior information of joint motion trajectory into temporal tokens. 
The KPA and TPA can enhance the capabilities of modeling global correlations in the self-attention mechanisms effectively.
% generate Q, K, V vectors with prior knowledge for the vanilla self-attention mechanisms, 
% which can enhance their capabilities of modeling global correlations more effectively.
Experimental results on three benchmarks demonstrate that our method is effective in improving the performance with only a very small increase in the computational overhead. Moreover, our KPA and TPA can be integrated into various transformer-based 3D pose estimators as lightweight plug-and-play modules.

\section*{Acknowledgments}
The work described in this paper is supported, in part, by the Innovation and Technology Commission of Hong Kong under grant ITP/028/21TP and by the Laboratory for Artificial Intelligence in Design (Project Code: RP1-1) under InnoHK Research Clusters, Hong Kong Special Administrative Region.
%\section*{acknowledgments}

{
    \small
    \bibliographystyle{ieeenat_fullname}
    \bibliography{main}

\begin{thebibliography}{48}
\providecommand{\natexlab}[1]{#1}
\providecommand{\url}[1]{\texttt{#1}}
\expandafter\ifx\csname urlstyle\endcsname\relax
  \providecommand{\doi}[1]{doi: #1}\else
  \providecommand{\doi}{doi: \begingroup \urlstyle{rm}\Url}\fi

\bibitem[Cai et~al.(2019)Cai, Ge, Liu, Cai, Cham, Yuan, and Thalmann]{cai2019exploiting}
Yujun Cai, Liuhao Ge, Jun Liu, Jianfei Cai, Tat-Jen Cham, Junsong Yuan, and Nadia~Magnenat Thalmann.
\newblock Exploiting spatial-temporal relationships for 3d pose estimation via graph convolutional networks.
\newblock In \emph{Proceedings of the IEEE/CVF international conference on computer vision}, pages 2272--2281, 2019.

\bibitem[Chen et~al.(2021)Chen, Fang, Shen, Zhu, Chen, and Luo]{chen2021anatomy}
Tianlang Chen, Chen Fang, Xiaohui Shen, Yiheng Zhu, Zhili Chen, and Jiebo Luo.
\newblock Anatomy-aware 3d human pose estimation with bone-based pose decomposition.
\newblock \emph{IEEE Transactions on Circuits and Systems for Video Technology}, 32\penalty0 (1):\penalty0 198--209, 2021.

\bibitem[Chen et~al.(2018)Chen, Wang, Peng, Zhang, Yu, and Sun]{chen2018cascaded}
Yilun Chen, Zhicheng Wang, Yuxiang Peng, Zhiqiang Zhang, Gang Yu, and Jian Sun.
\newblock Cascaded pyramid network for multi-person pose estimation.
\newblock In \emph{Proceedings of the IEEE conference on computer vision and pattern recognition}, pages 7103--7112, 2018.

\bibitem[Cheng et~al.(2021)Cheng, Wang, Yang, and Tan]{cheng2021graph}
Yu Cheng, Bo Wang, Bo Yang, and Robby~T Tan.
\newblock Graph and temporal convolutional networks for 3d multi-person pose estimation in monocular videos.
\newblock In \emph{Proceedings of the AAAI Conference on Artificial Intelligence}, pages 1157--1165, 2021.

\bibitem[Ci et~al.(2019)Ci, Wang, Ma, and Wang]{ci2019optimizing}
Hai Ci, Chunyu Wang, Xiaoxuan Ma, and Yizhou Wang.
\newblock Optimizing network structure for 3d human pose estimation.
\newblock In \emph{Proceedings of the IEEE/CVF international conference on computer vision}, pages 2262--2271, 2019.

\bibitem[Defferrard et~al.(2016)Defferrard, Bresson, and Vandergheynst]{defferrard2016convolutional}
Micha{\"e}l Defferrard, Xavier Bresson, and Pierre Vandergheynst.
\newblock Convolutional neural networks on graphs with fast localized spectral filtering.
\newblock \emph{Advances in neural information processing systems}, 29, 2016.

\bibitem[Gong et~al.(2023)Gong, Foo, Fan, Ke, Rahmani, and Liu]{gong2023diffpose}
Jia Gong, Lin~Geng Foo, Zhipeng Fan, Qiuhong Ke, Hossein Rahmani, and Jun Liu.
\newblock Diffpose: Toward more reliable 3d pose estimation.
\newblock In \emph{Proceedings of the IEEE/CVF Conference on Computer Vision and Pattern Recognition}, pages 13041--13051, 2023.

\bibitem[Gower(1975)]{gower1975generalized}
John~C Gower.
\newblock Generalized procrustes analysis.
\newblock \emph{Psychometrika}, 40:\penalty0 33--51, 1975.

\bibitem[Hagbi et~al.(2010)Hagbi, Bergig, El-Sana, and Billinghurst]{hagbi2010shape}
Nate Hagbi, Oriel Bergig, Jihad El-Sana, and Mark Billinghurst.
\newblock Shape recognition and pose estimation for mobile augmented reality.
\newblock \emph{IEEE transactions on visualization and computer graphics}, 17\penalty0 (10):\penalty0 1369--1379, 2010.

\bibitem[Hossain and Little(2018)]{hossain2018exploiting}
Mir Rayat~Imtiaz Hossain and James~J Little.
\newblock Exploiting temporal information for 3d human pose estimation.
\newblock In \emph{Proceedings of the European conference on computer vision (ECCV)}, pages 68--84, 2018.

\bibitem[Hu et~al.(2021)Hu, Zhang, Zhan, Zhang, and Wong]{hu2021conditional}
Wenbo Hu, Changgong Zhang, Fangneng Zhan, Lei Zhang, and Tien-Tsin Wong.
\newblock Conditional directed graph convolution for 3d human pose estimation.
\newblock In \emph{Proceedings of the 29th ACM International Conference on Multimedia}, pages 602--611, 2021.

\bibitem[Ionescu et~al.(2013)Ionescu, Papava, Olaru, and Sminchisescu]{ionescu2013human3}
Catalin Ionescu, Dragos Papava, Vlad Olaru, and Cristian Sminchisescu.
\newblock Human3. 6m: Large scale datasets and predictive methods for 3d human sensing in natural environments.
\newblock \emph{IEEE transactions on pattern analysis and machine intelligence}, 36\penalty0 (7):\penalty0 1325--1339, 2013.

\bibitem[Kisacanin et~al.(2005)Kisacanin, Pavlovic, and Huang]{kisacanin2005real}
Branislav Kisacanin, Vladimir Pavlovic, and Thomas~S Huang.
\newblock \emph{Real-time vision for human-computer interaction}.
\newblock Springer Science \& Business Media, 2005.

\bibitem[Li et~al.(2023)Li, Shi, Dai, Zheng, Wang, Sun, Guo, Li, Zou, and Xiong]{li2023pose}
Han Li, Bowen Shi, Wenrui Dai, Hongwei Zheng, Botao Wang, Yu Sun, Min Guo, Chenglin Li, Junni Zou, and Hongkai Xiong.
\newblock Pose-oriented transformer with uncertainty-guided refinement for 2d-to-3d human pose estimation.
\newblock In \emph{Proceedings of the AAAI Conference on Artificial Intelligence}, pages 1296--1304, 2023.

\bibitem[Li et~al.(2018)Li, Wang, Zhu, and Huang]{li2018adaptive}
Ruoyu Li, Sheng Wang, Feiyun Zhu, and Junzhou Huang.
\newblock Adaptive graph convolutional neural networks.
\newblock In \emph{Proceedings of the AAAI conference on artificial intelligence}, 2018.

\bibitem[Li and Chan(2015)]{li20153d}
Sijin Li and Antoni~B Chan.
\newblock 3d human pose estimation from monocular images with deep convolutional neural network.
\newblock In \emph{Computer Vision--ACCV 2014: 12th Asian Conference on Computer Vision, Singapore, Singapore, November 1-5, 2014, Revised Selected Papers, Part II 12}, pages 332--347. Springer, 2015.

\bibitem[Li et~al.(2022{\natexlab{a}})Li, Liu, Ding, Liu, Wang, and Yang]{li2022exploiting}
Wenhao Li, Hong Liu, Runwei Ding, Mengyuan Liu, Pichao Wang, and Wenming Yang.
\newblock Exploiting temporal contexts with strided transformer for 3d human pose estimation.
\newblock \emph{IEEE Transactions on Multimedia}, 25:\penalty0 1282--1293, 2022{\natexlab{a}}.

\bibitem[Li et~al.(2022{\natexlab{b}})Li, Liu, Tang, Wang, and Van~Gool]{li2022mhformer}
Wenhao Li, Hong Liu, Hao Tang, Pichao Wang, and Luc Van~Gool.
\newblock Mhformer: Multi-hypothesis transformer for 3d human pose estimation.
\newblock In \emph{Proceedings of the IEEE/CVF Conference on Computer Vision and Pattern Recognition}, pages 13147--13156, 2022{\natexlab{b}}.

\bibitem[Liu et~al.(2020{\natexlab{a}})Liu, Ding, Zou, Wang, and Tang]{liu2020comprehensive}
Kenkun Liu, Rongqi Ding, Zhiming Zou, Le Wang, and Wei Tang.
\newblock A comprehensive study of weight sharing in graph networks for 3d human pose estimation.
\newblock In \emph{Computer Vision--ECCV 2020: 16th European Conference, Glasgow, UK, August 23--28, 2020, Proceedings, Part X 16}, pages 318--334. Springer, 2020{\natexlab{a}}.

\bibitem[Liu and Yuan(2018)]{liu2018recognizing}
Mengyuan Liu and Junsong Yuan.
\newblock Recognizing human actions as the evolution of pose estimation maps.
\newblock In \emph{Proceedings of the IEEE conference on computer vision and pattern recognition}, pages 1159--1168, 2018.

\bibitem[Liu et~al.(2017)Liu, Liu, and Chen]{liu2017enhanced}
Mengyuan Liu, Hong Liu, and Chen Chen.
\newblock Enhanced skeleton visualization for view invariant human action recognition.
\newblock \emph{Pattern Recognition}, 68:\penalty0 346--362, 2017.

\bibitem[Liu et~al.(2020{\natexlab{b}})Liu, Shen, Wang, Chen, Cheung, and Asari]{liu2020attention}
Ruixu Liu, Ju Shen, He Wang, Chen Chen, Sen-ching Cheung, and Vijayan Asari.
\newblock Attention mechanism exploits temporal contexts: Real-time 3d human pose reconstruction.
\newblock In \emph{Proceedings of the IEEE/CVF Conference on Computer Vision and Pattern Recognition}, pages 5064--5073, 2020{\natexlab{b}}.

\bibitem[Martinez et~al.(2017)Martinez, Hossain, Romero, and Little]{martinez2017simple}
Julieta Martinez, Rayat Hossain, Javier Romero, and James~J Little.
\newblock A simple yet effective baseline for 3d human pose estimation.
\newblock In \emph{Proceedings of the IEEE international conference on computer vision}, pages 2640--2649, 2017.

\bibitem[Mehta et~al.(2017)Mehta, Rhodin, Casas, Fua, Sotnychenko, Xu, and Theobalt]{mehta2017monocular}
Dushyant Mehta, Helge Rhodin, Dan Casas, Pascal Fua, Oleksandr Sotnychenko, Weipeng Xu, and Christian Theobalt.
\newblock Monocular 3d human pose estimation in the wild using improved cnn supervision.
\newblock In \emph{2017 international conference on 3D vision (3DV)}, pages 506--516. IEEE, 2017.

\bibitem[Moon et~al.(2019)Moon, Chang, and Lee]{moon2019camera}
Gyeongsik Moon, Ju~Yong Chang, and Kyoung~Mu Lee.
\newblock Camera distance-aware top-down approach for 3d multi-person pose estimation from a single rgb image.
\newblock In \emph{Proceedings of the IEEE/CVF international conference on computer vision}, pages 10133--10142, 2019.

\bibitem[Newell et~al.(2016)Newell, Yang, and Deng]{newell2016stacked}
Alejandro Newell, Kaiyu Yang, and Jia Deng.
\newblock Stacked hourglass networks for human pose estimation.
\newblock In \emph{Computer Vision--ECCV 2016: 14th European Conference, Amsterdam, The Netherlands, October 11-14, 2016, Proceedings, Part VIII 14}, pages 483--499. Springer, 2016.

\bibitem[Park et~al.(2016)Park, Hwang, and Kwak]{park20163d}
Sungheon Park, Jihye Hwang, and Nojun Kwak.
\newblock 3d human pose estimation using convolutional neural networks with 2d pose information.
\newblock In \emph{Computer Vision--ECCV 2016 Workshops: Amsterdam, The Netherlands, October 8-10 and 15-16, 2016, Proceedings, Part III 14}, pages 156--169. Springer, 2016.

\bibitem[Pavlakos et~al.(2018)Pavlakos, Zhou, and Daniilidis]{pavlakos2018ordinal}
Georgios Pavlakos, Xiaowei Zhou, and Kostas Daniilidis.
\newblock Ordinal depth supervision for 3d human pose estimation.
\newblock In \emph{Proceedings of the IEEE conference on computer vision and pattern recognition}, pages 7307--7316, 2018.

\bibitem[Pavllo et~al.(2019)Pavllo, Feichtenhofer, Grangier, and Auli]{pavllo20193d}
Dario Pavllo, Christoph Feichtenhofer, David Grangier, and Michael Auli.
\newblock 3d human pose estimation in video with temporal convolutions and semi-supervised training.
\newblock In \emph{Proceedings of the IEEE/CVF conference on computer vision and pattern recognition}, pages 7753--7762, 2019.

\bibitem[Shan et~al.(2022)Shan, Liu, Zhang, Wang, Ma, and Gao]{shan2022p}
Wenkang Shan, Zhenhua Liu, Xinfeng Zhang, Shanshe Wang, Siwei Ma, and Wen Gao.
\newblock P-stmo: Pre-trained spatial temporal many-to-one model for 3d human pose estimation.
\newblock In \emph{European Conference on Computer Vision}, pages 461--478. Springer, 2022.

\bibitem[Shan et~al.(2023)Shan, Liu, Zhang, Wang, Han, Wang, Ma, and Gao]{shan2023diffusion}
Wenkang Shan, Zhenhua Liu, Xinfeng Zhang, Zhao Wang, Kai Han, Shanshe Wang, Siwei Ma, and Wen Gao.
\newblock Diffusion-based 3d human pose estimation with multi-hypothesis aggregation.
\newblock \emph{arXiv preprint arXiv:2303.11579}, 2023.

\bibitem[Sigal et~al.(2010)Sigal, Balan, and Black]{sigal2010humaneva}
Leonid Sigal, Alexandru~O Balan, and Michael~J Black.
\newblock Humaneva: Synchronized video and motion capture dataset and baseline algorithm for evaluation of articulated human motion.
\newblock \emph{International journal of computer vision}, 87\penalty0 (1-2):\penalty0 4--27, 2010.

\bibitem[Svenstrup et~al.(2009)Svenstrup, Tranberg, Andersen, and Bak]{svenstrup2009pose}
Mikael Svenstrup, Soren Tranberg, Hans~Jorgen Andersen, and Thomas Bak.
\newblock Pose estimation and adaptive robot behaviour for human-robot interaction.
\newblock In \emph{2009 IEEE International Conference on Robotics and Automation}, pages 3571--3576. IEEE, 2009.

\bibitem[Tang et~al.(2023)Tang, Qiu, Hao, Hong, and Yao]{tang20233d}
Zhenhua Tang, Zhaofan Qiu, Yanbin Hao, Richang Hong, and Ting Yao.
\newblock 3d human pose estimation with spatio-temporal criss-cross attention.
\newblock In \emph{Proceedings of the IEEE/CVF Conference on Computer Vision and Pattern Recognition}, pages 4790--4799, 2023.

\bibitem[Tekin et~al.(2016)Tekin, Rozantsev, Lepetit, and Fua]{tekin2016direct}
Bugra Tekin, Artem Rozantsev, Vincent Lepetit, and Pascal Fua.
\newblock Direct prediction of 3d body poses from motion compensated sequences.
\newblock In \emph{Proceedings of the IEEE Conference on Computer Vision and Pattern Recognition}, pages 991--1000, 2016.

\bibitem[Vaswani et~al.(2017)Vaswani, Shazeer, Parmar, Uszkoreit, Jones, Gomez, Kaiser, and Polosukhin]{vaswani2017attention}
Ashish Vaswani, Noam Shazeer, Niki Parmar, Jakob Uszkoreit, Llion Jones, Aidan~N Gomez, {\L}ukasz Kaiser, and Illia Polosukhin.
\newblock Attention is all you need.
\newblock \emph{Advances in neural information processing systems}, 30, 2017.

\bibitem[Wang et~al.(2020)Wang, Yan, Xiong, and Lin]{wang2020motion}
Jingbo Wang, Sijie Yan, Yuanjun Xiong, and Dahua Lin.
\newblock Motion guided 3d pose estimation from videos.
\newblock In \emph{European Conference on Computer Vision}, pages 764--780. Springer, 2020.

\bibitem[Wehrbein et~al.(2021)Wehrbein, Rudolph, Rosenhahn, and Wandt]{wehrbein2021probabilistic}
Tom Wehrbein, Marco Rudolph, Bodo Rosenhahn, and Bastian Wandt.
\newblock Probabilistic monocular 3d human pose estimation with normalizing flows.
\newblock In \emph{Proceedings of the IEEE/CVF international conference on computer vision}, pages 11199--11208, 2021.

\bibitem[Xu and Takano(2021)]{xu2021graph}
Tianhan Xu and Wataru Takano.
\newblock Graph stacked hourglass networks for 3d human pose estimation.
\newblock In \emph{Proceedings of the IEEE/CVF conference on computer vision and pattern recognition}, pages 16105--16114, 2021.

\bibitem[Yu et~al.(2023)Yu, Zhang, Liu, Zhong, Liu, and Chen]{yu2023gla}
Bruce~XB Yu, Zhi Zhang, Yongxu Liu, Sheng-hua Zhong, Yan Liu, and Chang~Wen Chen.
\newblock Gla-gcn: Global-local adaptive graph convolutional network for 3d human pose estimation from monocular video.
\newblock In \emph{Proceedings of the IEEE/CVF International Conference on Computer Vision}, pages 8818--8829, 2023.

\bibitem[Zhang et~al.(2022{\natexlab{a}})Zhang, Chen, and Tu]{zhang2022uncertainty}
Jinlu Zhang, Yujin Chen, and Zhigang Tu.
\newblock Uncertainty-aware 3d human pose estimation from monocular video.
\newblock In \emph{Proceedings of the 30th ACM International Conference on Multimedia}, pages 5102--5113, 2022{\natexlab{a}}.

\bibitem[Zhang et~al.(2022{\natexlab{b}})Zhang, Tu, Yang, Chen, and Yuan]{zhang2022mixste}
Jinlu Zhang, Zhigang Tu, Jianyu Yang, Yujin Chen, and Junsong Yuan.
\newblock Mixste: Seq2seq mixed spatio-temporal encoder for 3d human pose estimation in video.
\newblock In \emph{Proceedings of the IEEE/CVF conference on computer vision and pattern recognition}, pages 13232--13242, 2022{\natexlab{b}}.

\bibitem[Zhao et~al.(2019)Zhao, Peng, Tian, Kapadia, and Metaxas]{zhao2019semantic}
Long Zhao, Xi Peng, Yu Tian, Mubbasir Kapadia, and Dimitris~N Metaxas.
\newblock Semantic graph convolutional networks for 3d human pose regression.
\newblock In \emph{Proceedings of the IEEE/CVF conference on computer vision and pattern recognition}, pages 3425--3435, 2019.

\bibitem[Zhao et~al.(2023)Zhao, Zheng, Liu, Wang, and Chen]{zhao2023poseformerv2}
Qitao Zhao, Ce Zheng, Mengyuan Liu, Pichao Wang, and Chen Chen.
\newblock Poseformerv2: Exploring frequency domain for efficient and robust 3d human pose estimation.
\newblock In \emph{Proceedings of the IEEE/CVF Conference on Computer Vision and Pattern Recognition}, pages 8877--8886, 2023.

\bibitem[Zhao et~al.(2022)Zhao, Wang, and Tian]{zhao2022graformer}
Weixi Zhao, Weiqiang Wang, and Yunjie Tian.
\newblock Graformer: Graph-oriented transformer for 3d pose estimation.
\newblock In \emph{Proceedings of the IEEE/CVF Conference on Computer Vision and Pattern Recognition}, pages 20438--20447, 2022.

\bibitem[Zheng et~al.(2021)Zheng, Zhu, Mendieta, Yang, Chen, and Ding]{zheng20213d}
Ce Zheng, Sijie Zhu, Matias Mendieta, Taojiannan Yang, Chen Chen, and Zhengming Ding.
\newblock 3d human pose estimation with spatial and temporal transformers.
\newblock In \emph{Proceedings of the IEEE/CVF International Conference on Computer Vision}, pages 11656--11665, 2021.

\bibitem[Zhu et~al.(2021)Zhu, Xu, Shen, Ji, Gao, and Shen]{zhu2021posegtac}
Yiran Zhu, Xing Xu, Fumin Shen, Yanli Ji, Lianli Gao, and Heng~Tao Shen.
\newblock Posegtac: Graph transformer encoder-decoder with atrous convolution for 3d human pose estimation.
\newblock In \emph{IJCAI}, pages 1359--1365, 2021.

\bibitem[Zou and Tang(2021)]{zou2021modulated}
Zhiming Zou and Wei Tang.
\newblock Modulated graph convolutional network for 3d human pose estimation.
\newblock In \emph{Proceedings of the IEEE/CVF international conference on computer vision}, pages 11477--11487, 2021.

\end{thebibliography}
}


\end{document}